\documentclass[]{rtp_tech_report}

\usepackage{amsmath}
\usepackage{amssymb}
\usepackage{mathtools}

\usepackage{graphicx}
\usepackage{booktabs}
\usepackage{tabularx}
\usepackage{array}
\usepackage{wrapfig}
\usepackage{needspace}
\makeatletter
\newcommand{\finishablationwrap}{%
  \par
  \ifnum\c@WF@wrappedlines>1
    \@tempdima=\baselineskip
    \multiply\@tempdima by \numexpr\c@WF@wrappedlines-1\relax
    \vspace*{\@tempdima}%
  \fi
  \WFclear
}
\makeatother

\usepackage{xcolor}
\usepackage{colortbl}

\definecolor{oursorange}{HTML}{FFF4EB}
\definecolor{ablationblue}{RGB}{232,241,251}

\newcommand{\method}{\textbf{ADIndex}\xspace}

\title{ADIndex: Layer-Adaptive Sparse Attention Indexing via Group-Level Distillation}

\author{%
  \parbox{0.95\textwidth}{\centering
    Jin-Fei Qi$^{1,2,\sharp}$\hspace{1em}%
    Yiduo Li$^{2}$\hspace{1em}%
    Jiang-Xin Shi$^{1,\S}$\hspace{1em}%
    Kan Liu$^{2}$\\[0.45em]
    Tao Lan$^{2}$\hspace{1em}%
    Hanlin Tang$^{2,\dagger}$\hspace{1em}%
    Yu-Feng Li$^{1,\S}$
  }%
}

\affiliation[1]{Nanjing University}
\affiliation[2]{Alibaba Group}

\checkdata[E-mail]{\email{qijf@lamda.nju.edu.cn; shijx@lamda.nju.edu.cn; liyf@nju.edu.cn}}

\abstract{
    DeepSeek Sparse Attention (DSA) reduces the computational cost of attention through token-level selection. However, its indexer still scores the entire causal prefix for each query, making indexing a dominant source of latency in long-context inference. Notably, the indexer alone can account for more than 50\% of total prefill GPU time at long context lengths.
To address this bottleneck, we propose \method{}, an \textbf{A}daptive and \textbf{D}istillation-driven \textbf{Index}ing method for token-level sparse attention.
\method{} integrates three key components:
\textup{(\romannumeral1)} a layer-wise calibration strategy that adaptively accommodates cross-query retrieval overlap, motivated by our observation of pronounced inter-layer heterogeneity;
\textup{(\romannumeral2)} a group-level distillation scheme that leverages this heterogeneity to efficiently train adaptive layer-wise group indexers, capturing salient retrieval signals from individual queries; and
\textup{(\romannumeral3)} a low-latency inference strategy that reduces repeated full-prefix scans through token bank sharing within query groups while retaining query-specific refinement.
Experiments on DeepSeek-V3.2 and GLM-5.1 demonstrate that
\method{} preserves near-lossless task performance on LongBench,
LongBench v2, and RULER while delivering substantial efficiency
gains over standard DSA, including prefill and decoding kernel
speedups of up to $6.22\times$ and $3.40\times$, respectively,
at a context length of 256K across the layer-wise reuse
configurations, as well as an end-to-end serving speedup
of $1.54\times$ at a context length of 128K.

}

\begin{document}

\maketitle

\begingroup
\renewcommand{\thefootnote}{}
\footnotetext{%
  \textsuperscript{\ensuremath{\dagger}}Project lead
  \quad
  \textsuperscript{\S}Corresponding author
  \quad
  \textsuperscript{\ensuremath{\sharp}}Work done during internship at Alibaba
}
\endgroup

\section{Introduction}
\label{sec:introduction}

\begin{figure*}[t]
    \centering
    \includegraphics[width=\textwidth]{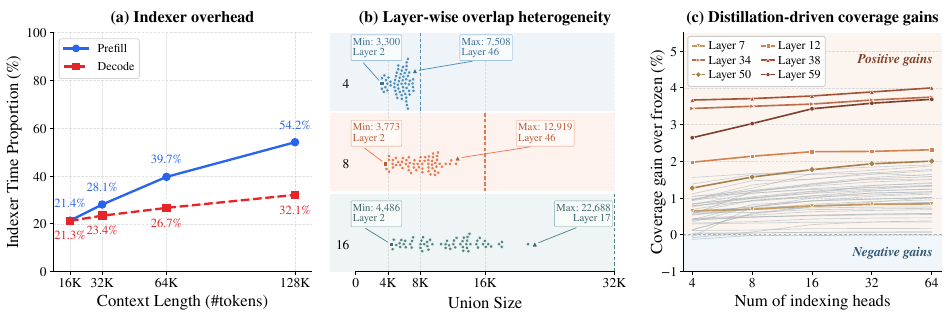}
    \caption{\textbf{Motivating observations for \method{}.}
    \textbf{(a) Indexer overhead.} The plot shows
    indexer time proportion during prefill and decoding
    at different context lengths. The indexer can alone account for 54.2\% of total prefill GPU time at 128k context length.
    \textbf{(b) Layer-wise overlap heterogeneity.}
    The plot shows per-layer maximum union sizes of the original indexer’s Top-$K$ selections ($K=2048$) for query group sizes $G\in\{4,8,16\}$, computed over all complete prefill query groups from 100 sequences of approximately 128K tokens each. Colored vertical dashed lines indicate the corresponding no-overlap upper bounds $GK$.
    \textbf{(c) Distillation-driven coverage gains.} The plot shows
    absolute gains in Top-2048 coverage of the distilled
group indexer over the frozen mean-pooling
baseline.
Coverage measures the fraction of the original DSA
indexer's per-query Top-2048 selections included
in the shared bank.
    The group size $G=16$ and bank budget
    $P=8\mathrm{K}$ are fixed, while the number of indexing
    heads in the group indexer varies.
    Gains are positive in almost all layers and remain
    robust across head counts.
    Selected layers from shallow (7, 12), mid (34, 38), and deep (50, 59)
parts of the model are highlighted in warm colors,
while the remaining layers are shown in gray.
    Pale warm and cool backgrounds indicate positive
    and negative gains, respectively; the horizontal
    dashed line marks zero gain.}
    \label{fig:introduction_overview}
\end{figure*}

Long-context capability has become a core requirement for modern large language models (LLMs), supporting applications such as repository-level code understanding, multi-document synthesis, and long-running agentic workflows~\citep{deepseekai2025deepseekv32,glm5team2026glm5,qiu2026qwen38next,qwenteam2026qwen38omni}.
However, the computational cost of full self-attention scales quadratically with sequence length, creating a major bottleneck for efficient long-context inference.
Sparse attention addresses this challenge by restricting
attention computation to a subset of relevant tokens
\citep{lu2025moba,gao2024seerattention,zhou2026rtpurbo,tang2025razorattention}.
Among these approaches, DeepSeek Sparse Attention (DSA)
has emerged as a trainable token-level sparse attention mechanism
adopted in production-scale LLMs, including
DeepSeek-V3.2 and GLM-5.1%
~\citep{deepseekai2025deepseekv32,glm5team2026glm5}.
DSA employs a lightweight token-wise indexer
to score all tokens within each query's causal prefix
and select the Top-$K$ tokens.
These selected tokens are then processed by
a downstream Sparse Multi-head Latent Attention
(Sparse MLA) operator, substantially reducing
attention computation while enabling fine-grained
token selection.

Although DSA reduces the cost of attention computation,
it still incurs substantial overhead in identifying
the tokens to select.
For each query, the indexer must score the entire
causal prefix to determine its Top-$K$ tokens.
This repeated full-prefix scanning leads to
$\mathcal{O}(L^2)$ indexing complexity during prefill
and $\mathcal{O}(L)$ complexity per query during
decoding, where $L$ denotes the context length.
As shown in Figure~\ref{fig:introduction_overview}(a),
the indexer introduces substantial overhead in both stages,
with its relative contribution increasing steadily with context length.
At a context length of 128K, the indexer alone accounts
for 54.2\% of total prefill GPU time, making it a key
constraint on long-context inference efficiency.

Recent studies have accelerated DSA indexing through
hierarchical filtering (HISA), query-dependent head
routing (MISA), cross-layer index reuse (IndexCache),
and cross-query candidate sharing (PIVOT)%
~\citep{xu2026hisa,zhou2026misa,bai2026indexcache,liu2026pivot}.
Despite these advances, 
our analysis reveals
two further observations.
First, we observe pronounced {\rmfamily\bfseries\itshape layer-wise overlap
heterogeneity}: even with fixed $G$ and $K$, the degree
of overlap among neighboring queries' Top-$K$ selections
varies substantially across attention layers.
As shown in Figure~\ref{fig:introduction_overview}(b),
for $G=16$ and $K=2048$, the per-layer maximum union
sizes range from 4,486 to 22,688 on 128K inputs,
all below the no-overlap upper bound of $GK=32{,}768$.
These results reveal substantial cross-query overlap
while highlighting layer-dependent capacity
requirements for shared token banks.
This heterogeneity motivates adapting both query group
sizes and shared bank budgets to individual layers.
Second, we observe {\rmfamily\bfseries\itshape distillation-driven coverage
gains}: lightweight distillation enables the shared bank to cover a larger fraction of each query's Top-$K$ tokens selected
by the original DSA indexer.
For each query, scanning only the shared bank with the original indexer exactly recovers the full-prefix Top-$K$ tokens, provided that the bank contains all $K$ of them.
As shown in Figure~\ref{fig:introduction_overview}(c),
the distilled lightweight group indexer achieves higher
Top-$K$ coverage than its frozen counterpart in almost
all layers, with gains remaining robust across different indexing-head counts.

Motivated by these observations, we introduce \method{}, a layer-adaptive and distillation-driven indexing method for token-level sparse attention.
First, to address layer-wise overlap heterogeneity, we perform offline calibration to select query group sizes and shared token bank budgets for individual layers, enabling flexible candidate sharing tailored to each layer's cross-query overlap patterns.
Second, we train a lightweight group indexer through
efficient distillation from the original DSA indexer.
The distillation scheme improves Top-$K$ token coverage within each
layer's shared token bank budget at low cost, with gains that remain
robust as the number of indexing heads is reduced.
During inference, the lightweight group indexer
constructs a shared token bank with a single
full-prefix scan per query group, followed by
query-specific refinement.
This design substantially reduces full-prefix scanning overhead while preserving query-specific token selection. Experiments on long-context benchmarks show that \method{}
preserves near-lossless performance while delivering substantial efficiency gains.

Our main contributions are summarized as follows:
\begin{itemize}
    \item
    We characterize substantial layer-wise heterogeneity
    in cross-query retrieval overlap and propose a
    layer-adaptive calibration strategy.
    By adaptively selecting query group sizes and
    shared token bank budgets, this strategy enables
    flexible sharing tailored to each
    layer's cross-query overlap patterns.
    \item
    We present \method{}, an adaptive and efficient
indexing method for token-level sparse attention.
Through group-level distillation, a
shared token bank is constructed to improve  Top-$K$ coverage
even with fewer indexing heads. At inference time, a low-latency strategy shares token banks across neighboring queries
to reduce redundant full-prefix scanning while
preserving fine-grained token selection.

    \item
    Our evaluation covers DeepSeek-V3.2 and GLM-5.1
on three long-context benchmarks: LongBench,
LongBench v2, and RULER.
Among the layer-wise reuse configurations, \method{} achieves
prefill and decoding kernel speedups of up to $6.22\times$
and $3.40\times$, respectively, at a context length of 256K
relative to standard DSA.
It further delivers an end-to-end serving speedup of $1.54\times$,
while maintaining near-lossless task performance.
\end{itemize}

\section{Related Work}
\label{sec:related_work}

\paragraph{Sparse Attention.}
Sparse attention reduces the computational cost of
long-context modeling by sparsifying query--key interactions.
Early approaches employ predefined sparsity patterns
\citep{child2019generating,beltagy2020longformer,zaheer2020bigbird}.
More recent methods explore content-adaptive sparsity.
Training-free approaches, such as MInference, Quest,
and FlexPrefill, identify input-dependent sparse patterns
or estimate token-block relevance during inference
\citep{jiang2024minference,tang2024quest,lai2025flexprefill}.
Other approaches incorporate sparse attention during
model training or adapt lightweight selection modules
after pretraining.
SeerAttention trains learnable gates through distillation
from pooled full-attention maps to predict block-level
sparsity \citep{gao2024seerattention}.
MoBA selects token blocks using query-dependent routing
based on mean-pooled keys \citep{lu2025moba},
while Native Sparse Attention (NSA) combines compressed,
selective, and sliding-window attention in a natively
trainable architecture \citep{yuan2025nsa}.
Selecting entire blocks facilitates efficient computation
but limits selection flexibility at the individual-token level.
At a finer granularity, RTPurbo enables token-level sparse
inference in pretrained dense-attention models through
a lightweight token indexer and limited additional training
\citep{zhou2026rtpurbo}.
Token-level indexers provide more flexible selection,
but their efficiency also depends on the cost of
identifying relevant tokens.

\paragraph{DSA Indexer Acceleration.}
DeepSeek Sparse Attention
\citep{deepseekai2025deepseekv32}
achieves fine-grained token selection through
a learned indexer.
However, the indexer still scores the full
prefix for every query, resulting in
quadratic indexing complexity during prefill.
Recent methods address this overhead by
reducing the cost of individual indexing
operations or exploiting redundancy across them.
HISA \citep{xu2026hisa} employs hierarchical
block-level filtering followed by fine-grained
token scoring to reduce the search space.
MISA \citep{zhou2026misa} reduces head-wise
computation by dynamically activating a
query-dependent subset of indexer heads.
IndexCache \citep{bai2026indexcache} exploits
cross-layer redundancy by reusing Top-$K$
indices across different layers.
PIVOT \citep{liu2026pivot} exploits
redundancy among neighboring queries through
training-free query aggregation, sharing
a full-prefix scan across each group.

\section{Preliminaries}
\label{sec:preliminaries}

DeepSeek Sparse Attention (DSA) decouples
token selection from attention computation
through two components: a lightweight
token-wise indexer and a Sparse MLA operator.

\paragraph{DSA Indexer.}
For a query at position $t$ and a causally
visible position $s \le t$ in layer $\ell$,
the indexer computes a relevance score
using $H^I$ indexing heads:
\begin{equation}
I_{\ell,t,s}
=
\sum_{j=1}^{H^I}
w_{\ell,t,j}^I
\operatorname{ReLU}
\left(
\mathbf{q}_{\ell,t,j}^I
\cdot
\mathbf{k}_{\ell,s}^I
\right),
\label{eq:dsa_indexer_score}
\end{equation}
where $\mathbf{q}_{\ell,t,j}^I$ and
$w_{\ell,t,j}^I$ denote the indexing query
and gating weight of head $j$, respectively,
and $\mathbf{k}_{\ell,s}^I$ denotes the
indexing key at position $s$.

The indexer selects the $K$ highest-scoring
positions from the causal prefix:
\begin{equation}
\mathcal{T}_{\ell,t}
=
\operatorname{TopK}_{s\le t}
\left(I_{\ell,t,s},K\right),
\label{eq:dsa_topk}
\end{equation}
where $K=2048$ in our experiments.
Although the indexer uses lightweight
representations, it must score the entire
causal prefix independently for each query,
resulting in $\mathcal{O}(L^2)$ indexing
complexity during prefill and $\mathcal{O}(L)$ complexity per query during
decoding.

\paragraph{Sparse MLA.}
Given the selected index set
$\mathcal{T}_{\ell,t}$,
Sparse MLA computes attention only over
the corresponding latent key-value entries.
Let $\mathbf{c}_{\ell,s}^{\mathrm{KV}}$ denote the
latent key-value representation at
position $s$ in layer $\ell$.
The sparse attention computation is
expressed as
\begin{equation}
\mathbf{o}_{\ell,t}
=
\operatorname{SparseMLA}
\left(
\mathbf{h}_{\ell,t},
\{\mathbf{c}_{\ell,s}^{\mathrm{KV}}\}_{s\in\mathcal{T}_{\ell,t}}
\right),
\label{eq:sparse_mla}
\end{equation}
where $\mathbf{h}_{\ell,t}$ denotes the
query-side hidden representation.
By restricting attention to $K$ selected
tokens per query, Sparse MLA reduces
the attention computation complexity
from $\mathcal{O}(L^2)$ to
$\mathcal{O}(LK)$ for a sequence
of length $L$.

\section{Method}
\label{sec:method}

\begin{figure*}[t]
    \centering
    \includegraphics[width=\textwidth]{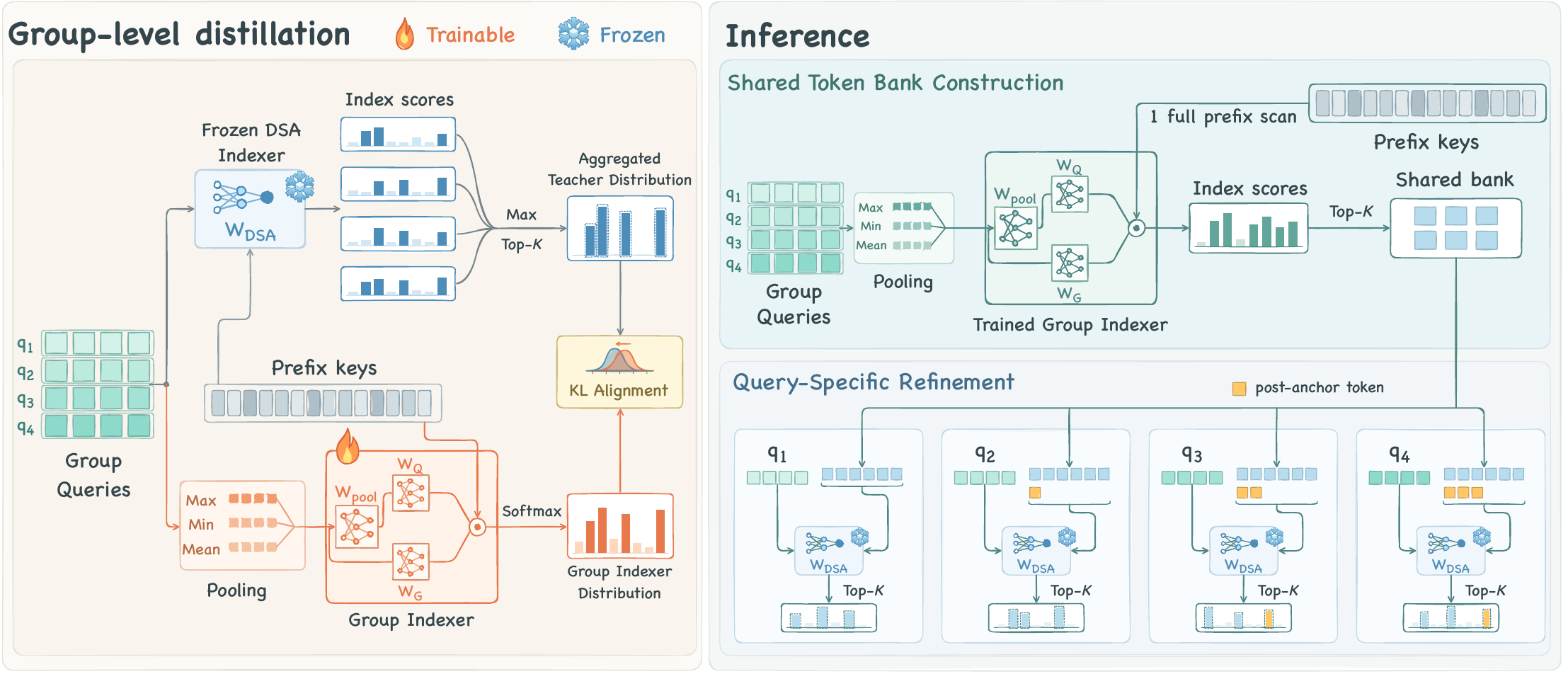}
    \caption{
    Overview of \method{}.
    During training, a lightweight group indexer
    learns the group-level retrieval requirements
    of neighboring queries through distillation
    from the frozen DSA indexer.
    During inference, the trained group indexer
    constructs a shared token bank through
    a single full-prefix scan per group.
    The original DSA indexer then performs
    query-specific Top-$K$ refinement over the
    shared bank augmented with newly visible post-anchor tokens.
}
    \label{fig:method_sketch}
\end{figure*}

We introduce \method{}, an adaptive and lightweight indexing
method for DSA that reduces redundant full-prefix scans
through distillation-driven token bank sharing across neighboring queries.
Figure~\ref{fig:method_sketch} illustrates the overall framework
of our method. First, to exploit the pronounced inter-layer overlap heterogeneity, we
employ a layer-wise reuse calibration strategy. We then introduce a
group-level distillation scheme that leverages this heterogeneity
to efficiently train adaptive, layer-specific group indexer.
Differently from the original DSA, we adopt a low-latency inference strategy that reuses
token banks within query groups, reducing repeated full-prefix
scans while retaining query-specific refinement.

\subsection{Layer-wise Reuse Calibration}
\label{sec:reuse_setting}

\paragraph{Token Bank Capacity Estimation.}
To account for the layer-wise overlap heterogeneity,
we determine the token bank capacity required for each
layer using calibration data.
Let $\mathcal{T}_{\ell,i}$ denote the Top-$K$
index set selected by the original DSA indexer
for query $i$ at layer $\ell$.
For each candidate group size $G$, we compute
the maximum union size of these index sets
across consecutive query groups:
\begin{equation}
U_{\ell,G}^{\max}
=
\max_{t\in\mathcal{A}_G}
\left|
\bigcup_{r=0}^{G-1}
\mathcal{T}_{\ell,t+r}
\right|,
\label{eq:calibration_max_union}
\end{equation}
where $\mathcal{A}_G$ denotes the set of valid
group anchors in the calibration dataset.
This statistic provides a conservative estimate
of the bank capacity required to
accommodate the original Top-$K$ selections
of neighboring queries.

\paragraph{Configuration Selection.}
A larger group size $G$ reduces redundant
full-prefix scanning by allowing more queries
to share a single scan, while a smaller
token bank budget $P$ reduces the cost of
query-specific refinement.
We define $S(G,P;L)$ as the joint indexer scoring
speedup for prefill of length $L$ and the fixed Multi-Token Prediction (MTP)
target verification workload specified in
Appendix~\ref{app:indexer_complexity}.
It is computed as the ratio of the total baseline
scoring count to the total scoring count of our
method across both stages. Each score evaluation
is assigned the same unit cost, irrespective of
the number of indexing heads.
For each layer $\ell$, we select the configuration
that maximizes the indexing speedup
at $L=128\mathrm{K}$, subject to the estimated
capacity constraint:
\begin{equation}
(G_\ell^*,P_\ell^*)
=
\operatorname*{arg\,max}_{
\substack{
G\in\{4,8,16\}\\
P\in\{4\mathrm{K},8\mathrm{K},16\mathrm{K}\}\\
U_{\ell,G}^{\max}\le P
}}
S(G,P;128\mathrm{K}).
\label{eq:calibration_optimal_setting}
\end{equation}
The indexing speedups for different
configurations are summarized in
Table~\ref{tab:app_joint_theoretical_speedup},
with detailed derivations provided in
Appendix~\ref{app:indexer_complexity}. For simplicity, we use
$G_\ell$ and $P_\ell$ to denote the calibrated reuse configuration for layer $\ell$.

\subsection{Group-level Distillation}
\label{sec:training}
We supervise the group indexer through group-level
distillation from the frozen DSA indexer.
Specifically, we aggregate the indexer's per-query
distributions into a group-level teacher distribution
and align the group indexer distribution
with this target.

\paragraph{Group Indexer Distribution.}
Consider a group of $G_\ell$ consecutive queries
$\mathcal{G}_t=\{t,\ldots,t+G_\ell-1\}$ starting
at position $t$ for layer $\ell$.
For each $i\in\mathcal{G}_t$, let $\mathbf{c}_i^{\mathrm{Q}}$
and $\mathbf{h}_i$ denote the query latent and
normalized hidden representation obtained
from the frozen backbone, respectively.
We aggregate the query latents using element-wise
max--min--mean pooling, while averaging
the hidden representations:
\begin{equation}
\bar{\mathbf{c}}_t^{\mathrm{Q}} = \mathbf{W}_{\mathrm{pool}}
[\mathbf{c}_t^{\mathrm{Q},\max};
 \mathbf{c}_t^{\mathrm{Q},\min};
 \mathbf{c}_t^{\mathrm{Q},\mathrm{mean}}],
\quad
\bar{\mathbf{h}}_t = \frac{1}{G_\ell}\sum_{i\in\mathcal{G}_t}\mathbf{h}_i,
\label{eq:group_student_representation}
\end{equation}
where $\mathbf{c}_t^{\mathrm{Q},\max}$,
$\mathbf{c}_t^{\mathrm{Q},\min}$, and
$\mathbf{c}_t^{\mathrm{Q},\mathrm{mean}}$ denote the
corresponding pooling results, and
$\mathbf{W}_{\mathrm{pool}}$ is a learnable
projection.

The aggregated representations are projected
into group-level indexing queries and head gates:
\begin{equation}
\bar{\mathbf{q}}_{t,j}^I =
\operatorname{RoPE}_t(
[\mathbf{W}_Q^S\bar{\mathbf{c}}_t^{\mathrm{Q}}]_j),
\quad
\bar w_{t,j}^I =
\sigma([\mathbf{W}_G^S\bar{\mathbf{h}}_t]_j),
\label{eq:student_query_gate}
\end{equation}
where $\mathbf{W}_Q^S$ and $\mathbf{W}_G^S$
are learnable projections, $[\cdot]_j$
denotes the output corresponding to head $j$,
and $\operatorname{RoPE}_t$ applies rotary
position embeddings at the group anchor $t$.
The group indexer employs $H_S^I$ indexing heads,
fewer than those of the original DSA indexer.
Using the frozen indexing keys $\mathbf{k}_s^I$,
the group indexer computes indexing scores:
\begin{equation}
\bar I_{t,s}
=\sum_{j=1}^{H_S^I}\bar w_{t,j}^I
\operatorname{ReLU}\!\left(
\bar{\mathbf{q}}_{t,j}^I
\cdot\mathbf{k}_s^I\right),
\quad s\le t.
\label{eq:group_student_score}
\end{equation}

We obtain the group indexer distribution by
applying a temperature-scaled softmax over
the anchor-visible prefix:
\begin{equation}
p_{t,s}^S
=\frac{\exp(\bar I_{t,s}/\tau_S)}
{\sum_{u\le t}\exp(\bar I_{t,u}/\tau_S)},
\quad s\le t,
\label{eq:group_student_distribution}
\end{equation}
where $\tau_S$ is the student temperature.

\paragraph{Aggregated Teacher Distribution.}
For each query $i\in\mathcal{G}_t$, we restrict
the frozen DSA indexer's scores $I_{i,s}$
to the anchor-visible prefix $s\le t$,
matching the positions scored by the group indexer.
We first standardize the restricted scores
and apply temperature scaling:
$z_{i,s}=(I_{i,s}-\mu_i)/(\sigma_i\tau_T)$,
where $\mu_i$ and $\sigma_i$ denote the mean
and standard deviation of the restricted scores,
respectively, and $\tau_T$ is the teacher temperature.
We then apply softmax to obtain per-query
distributions and aggregate them
using an element-wise maximum:
\begin{equation}
\pi_{i,s} = \frac{\exp(z_{i,s})}{\sum_{u\le t}\exp(z_{i,u})},
\quad
m_{t,s} = \max_{i\in\mathcal{G}_t}\pi_{i,s},
\quad s\le t.
\label{eq:group_teacher_aggregation}
\end{equation}
This aggregation emphasizes tokens that
are highly relevant to at least one query
in the group.

We select the $P_\ell$ highest-scoring positions according to
$m_{t,s}$ to form the teacher support
$\mathcal{S}_t$, where $P_\ell$ is the shared bank
budget selected by layer-wise reuse calibration.
The aggregated teacher distribution is obtained
by renormalizing $m_{t,s}$ over this support:
\begin{equation}
p_{t,s}^T=
\begin{cases}
\displaystyle
\frac{m_{t,s}}{\sum_{u\in\mathcal{S}_t}m_{t,u}},
& s\in\mathcal{S}_t,\\[6pt]
0, & \text{otherwise}.
\end{cases}
\label{eq:group_teacher_distribution}
\end{equation}

\paragraph{Distribution Alignment.}
We align the group indexer distribution $p_t^S$
with the aggregated teacher distribution $p_t^T$
by minimizing the forward Kullback--Leibler
(KL) divergence:
\begin{equation}
\mathcal{L}_t
=D_{\mathrm{KL}}(p_t^T\Vert p_t^S)
=\sum_{s\in\mathcal{S}_t}p_{t,s}^T
\log\frac{p_{t,s}^T}{p_{t,s}^S}.
\label{eq:group_distillation_loss}
\end{equation}
We optimize only the layer-specific group indexer
parameters $\mathbf{W}_{\mathrm{pool}}$,
$\mathbf{W}_Q^S$, and $\mathbf{W}_G^S$,
while keeping the pretrained backbone and
original DSA indexer frozen.

\subsection{Inference}
\label{sec:inference}

During inference, each layer adopts its calibrated
group size $G_\ell$ and shared token bank budget
$P_\ell$.
When reuse is enabled, \method{} performs
shared bank construction followed
by query-specific refinement.

\paragraph{Shared Token Bank Construction.}
For layer $\ell$, consider a query group
$\mathcal{G}_t=\{t,\ldots,t+G_\ell-1\}$
anchored at position $t$.
We use the trained group indexer to construct the shared bank only when the
anchor-visible prefix length $t+1$ exceeds
the bank budget $P_\ell$.
Otherwise, we fall back to the original DSA indexer.
When reuse is enabled, the trained group indexer
performs a single full-prefix scan to compute
group-level indexing scores $\bar I_{t,s}$
for $s\le t$, following
Eq.~\eqref{eq:group_student_score}.
It then selects the $P_\ell$ highest-scoring tokens to
construct a shared token bank:
\begin{equation}
\mathcal{P}_t
=
\operatorname{TopK}_{s\le t}
\left(\bar I_{t,s},P_\ell\right).
\label{eq:inference_pool_topk}
\end{equation}

\paragraph{Query-Specific Refinement.}

For each query $i\in\mathcal{G}_t$, we augment
the shared bank $\mathcal{P}_t$
with newly visible post-anchor tokens,
forming the candidate set
$\mathcal{C}_i=\mathcal{P}_t\cup\{t+1,\ldots,i\}$.
The original DSA indexer then computes
query-specific indexing scores $I_{i,s}$
over $\mathcal{C}_i$ following
Eq.~\eqref{eq:dsa_indexer_score},
and selects the final Top-$K$ indices:
\begin{equation}
\mathcal{T}_i
=
\operatorname{TopK}_{s\in\mathcal{C}_i}
\left(I_{i,s},K\right).
\label{eq:inference_final_topk}
\end{equation}

\paragraph{Application to Prefill and MTP.}
During prefill, we partition the input sequence
into consecutive, non-overlapping query groups
of size $G_\ell$ for layer $\ell$.
Eligible groups perform shared token bank
construction followed by query-specific
refinement, while incomplete groups fall back
to the original DSA indexer.
During MTP target verification, each verification
window processes 16 query tokens in parallel.
We partition these tokens into consecutive,
non-overlapping groups of size $G\in\{4,8,16\}$.
Reuse eligibility is determined once per
verification window based on its initial visible
prefix length.
If this length exceeds $P$, each group applies
the same two-stage procedure as in
prefill; otherwise, the entire window falls
back to the original DSA indexer.

\section{Experiments}
\label{sec:experiments}

We evaluate \method{} in terms of long-context
task performance and inference efficiency.
We compare it with the original DSA and existing
indexer acceleration methods on three long-context
benchmarks, measure kernel and end-to-end
inference speedups, and conduct ablation studies
to assess its key design choices.

\subsection{Experimental Settings}
\label{sec:experimental_settings}

\paragraph{Models and Benchmarks.}
We evaluate \method{} on DeepSeek-V3.2~\citep{deepseekai2025deepseekv32} and
GLM-5.1~\citep{glm5team2026glm5}, both in non-thinking mode.
We use LongBench~\citep{bai2024longbench} to evaluate
diverse long-context tasks, LongBench v2~\citep{bai2025longbenchv2}
to assess long-context understanding and reasoning
on challenging real-world tasks, and
RULER~\citep{hsieh2024ruler} to evaluate retrieval,
multi-hop tracing, information aggregation,
and question answering at controlled context lengths.
We evaluate RULER at 16K, 32K, 64K, and 128K tokens.
Additional benchmark details are provided in
Appendix~\ref{app:benchmark details}.

\paragraph{Baselines.}
We compare \method{} with the original DSA
and four existing DSA indexer acceleration methods:
HISA, MISA, IndexCache, and PIVOT.
Details of our baseline implementations are provided
in Appendix~\ref{app:baseline details}.

\paragraph{Calibration and Training.}
We use a corpus of 128K-token sequences constructed
from FineWeb-Edu and Cosmopedia V2 for offline
layer-wise reuse calibration and group indexer training.
We use 100 sequences for calibration and a disjoint
set of 1,000 sequences for training.
Layer-wise reuse configurations are determined
using the maximum-union criterion described in
Section~\ref{sec:reuse_setting}.
During training, the pretrained backbone and
original DSA indexer remain frozen, and only
the layer-specific group indexer is optimized.
Further details on data preparation and training
are provided in
Appendices~\ref{app:training data details}
and~\ref{app:group indexer training details}, respectively.

\paragraph{Implementation Details.}
We implement \method{} in SGLang~\citep{zheng2024sglang}
and conduct model evaluations on NVIDIA B300 GPUs.
Unless otherwise specified, the group indexer follows
the pooling strategy described in
Section~\ref{sec:training}, with 8 and 4 indexing
heads per layer for DeepSeek-V3.2 and GLM-5.1,
respectively.
The original DSA indexer performs query-specific
refinement with a final retrieval budget of $K=2048$.
We apply \method{} during both prefill and MTP target
verification, with 16 query tokens per verification
window.
Reuse activation and fallback follow
Section~\ref{sec:inference}.
The complete layer-wise reuse configurations and
additional implementation details are provided in
Appendix~\ref{app:layer-wise reuse configurations}.

\subsection{Main Results}
\label{sec:main_results}

\begin{table*}[t]
    \centering
    \small
    \setlength{\tabcolsep}{2pt}
    \renewcommand{\arraystretch}{1.15}

    \captionsetup{position=top}
    \caption{
    \textbf{Results (\%) on LongBench, LongBench v2, and RULER.}
    We report results for DeepSeek-V3.2 and GLM-5.1.
    LongBench categories include Single-Document QA (SQA),
    Multi-Document QA (MQA), Summarization (Sum),
    Few-Shot Learning (FS), Synthetic (Syn), and
    Code Completion (Code).
    LongBench v2 results are reported for the Easy and Hard
    difficulty subsets and the Short, Medium, and Long
    context-length subsets.
    RULER results are reported at context lengths of
    16K, 32K, 64K, and 128K.
    Avg denotes the mean score across the six task
    categories for LongBench, the overall accuracy
    for LongBench v2, and the mean score across the
    four context lengths for RULER.
    The best and second-best
    results are highlighted in bold and underlined,
    respectively.
    Orange-shaded rows indicate \method{}.
}
    \label{tab:longbench_longbenchv2_ruler_main_results}

    \renewcommand{\textbf}[1]{{\bfseries\sffamily #1}}
    \resizebox{\linewidth}{!}{%
        \begin{tabular}{l*{7}{c}|*{6}{c}|*{5}{c}}
            \toprule
            & \multicolumn{7}{c}{\textbf{LongBench}}
            & \multicolumn{6}{c}{\textbf{LongBench v2}}
            & \multicolumn{5}{c}{\textbf{RULER}} \\
            \cmidrule(lr){2-8}
            \cmidrule(lr){9-14}
            \cmidrule(lr){15-19}

            \textbf{Method}
            & \textbf{SQA}
            & \textbf{MQA}
            & \textbf{Sum}
            & \textbf{FS}
            & \textbf{Syn}
            & \textbf{Code}
            & \textbf{Avg}
            & \textbf{Easy}
            & \textbf{Hard}
            & \textbf{Short}
            & \textbf{Medium}
            & \textbf{Long}
            & \textbf{Avg}
            & \textbf{16K}
            & \textbf{32K}
            & \textbf{64K}
            & \textbf{128K}
            & \textbf{Avg} \\

            \midrule
            \multicolumn{19}{c}{\textbf{DeepSeek-V3.2}} \\
            \midrule

            DSA
            & 51.38 & \underline{53.89} & 22.48 & 67.95 & 67.33 & 73.40 & 56.07
            & 55.21 & \textbf{49.84} & 55.56 & \textbf{47.91} & 53.70& \underline{51.89}
            & \underline{96.72} & \underline{95.78} & \textbf{95.28} & \textbf{93.69} & \textbf{95.37} \\

            HISA
            & 51.74 & 53.81 & 22.23 & 67.82 & \textbf{68.83} & \textbf{74.38} & \textbf{56.47}
            & 51.04 & \underline{48.55} & 57.22 & 41.86 & 51.85 & 49.50
            & 94.34 & 85.45 & 69.03 & 47.91 & 74.18 \\

            MISA
            & 51.30 & 53.27 & \underline{22.56} & \underline{68.17} & 58.50 & \underline{74.32} & 54.69
            & 45.31 & 45.34 & 52.22 & 40.93 & 42.59 & 45.33
            & 96.29 & 95.08 & 93.95 & 84.93 & 92.56 \\

            IndexCache
            & 51.78 & 52.87 & \textbf{22.57} & 68.03 & 63.17 & 72.84 & 55.21
            & 52.60 & 47.91 & \underline{58.33} & 44.19 & 46.30 & 49.70
            & 96.45 & 95.76 & 93.33 & 90.13 & 93.92 \\

            PIVOT
            & \textbf{51.94} & 53.66 & 22.48 & 68.13 & 67.00 & 74.05 & 56.21
            & \underline{57.29} & 48.23 & 56.11 & 46.05 & \textbf{55.56} & 51.69
            & 96.47 & 95.66 & 94.51 & 92.18 & 94.71 \\

            \cmidrule(lr){1-19}
            \rowcolor{oursorange}
            \textbf{\method{}}
            & \underline{51.84} & \textbf{54.33} & 22.54 & \textbf{68.24} & \underline{67.50} & 72.90 & \underline{56.23}
            & \textbf{59.38} & \underline{48.55} & \textbf{58.89} & \underline{46.51} & \underline{54.63} & \textbf{52.68}
            & \textbf{96.81} & \textbf{95.81} & \underline{95.16} & \underline{93.34} & \underline{95.28} \\

            \midrule
            \multicolumn{19}{c}{\textbf{GLM-5.1}} \\
            \midrule

            DSA
            & 50.22 & 55.74 & \textbf{23.95} & 68.66 & 67.10 & 70.92 & 56.10
            & \underline{58.85} & \underline{50.80} & \underline{61.67} & \textbf{47.44} & \textbf{53.70} & \textbf{53.88}
            & 95.66 & \textbf{95.61} & \textbf{95.54} & \textbf{94.97} & \textbf{95.45} \\

            HISA
            & 50.10 & 54.91 & 23.86 & 68.47 & \underline{67.58} & \textbf{71.22} & 56.02
            & 57.81 & 49.84 & \underline{61.67} & \textbf{47.44} & 49.07 & 52.88
            & 94.08 & 87.68 & 79.96 & 64.04 & 81.44 \\

            MISA
            &50.22 &55.75 &23.76 &\textbf{69.29} &\textbf{68.39} &69.90 &\textbf{56.22}
            & 52.60&44.69 &60.56 &39.53 & 42.59&47.71
            &95.69 &94.54 & 93.50&86.39 &92.53 \\

            IndexCache
            & 49.87 & \underline{56.42} & 23.90 & \underline{68.72} & 66.56 & 69.42 & 55.82
            & \textbf{59.90} & 49.52 & 61.11 & \textbf{47.44} & \underline{52.78} & 53.48
            & 95.82 & 95.14 & 94.78 & 92.90 & 94.66 \\

            PIVOT
            & \textbf{50.42} & \textbf{56.73} & \underline{23.93} & 68.34 & \underline{67.58} & 68.99 & 56.00
            & \underline{58.85} & 49.52 & \textbf{64.44} & \underline{46.51} & 47.22 & 53.08
            & \textbf{96.04} & 95.45 & 95.05 & 93.80 & 95.09 \\

            \cmidrule(lr){1-19}
            \rowcolor{oursorange}
            \textbf{\method{}}
            & \underline{50.31} & 56.08 & 23.92 & 68.55 & 67.25 & \underline{70.94} & \underline{56.18}
            & 57.29 & \textbf{51.45} & \textbf{64.44} & 46.05 & 50.93 & \underline{53.68}
            & \underline{95.84} & \underline{95.57} & \underline{95.51} & \underline{94.49} & \underline{95.35} \\

            \bottomrule
        \end{tabular}%
    }
\end{table*}

\subsubsection{Task Performance}
\label{sec:task_performance}

Table~\ref{tab:longbench_longbenchv2_ruler_main_results}
compares \method{} with the original DSA and four
existing DSA indexer acceleration methods on
LongBench, LongBench v2, and RULER.

\paragraph{LongBench.}
\method{} achieves average scores of 56.23 and
56.18 on DeepSeek-V3.2 and GLM-5.1, respectively,
compared with 56.07 and 56.10 for the original DSA.
It ranks second among the evaluated methods on
both models, behind HISA and MISA, respectively.
These results demonstrate that \method{} maintains
comparable performance to the original DSA
across diverse long-context tasks.

\paragraph{LongBench v2.}
We further evaluate \method{} on LongBench v2,
which assesses long-context understanding and reasoning
across different difficulty levels and input lengths.
\method{} achieves overall accuracies of 52.68
and 53.68 on DeepSeek-V3.2 and GLM-5.1,
respectively, compared with 51.89 and 53.88
for the original DSA.
It achieves the highest overall accuracy among
the evaluated indexer acceleration methods
on both models.
Despite variations across difficulty and
context-length subsets, \method{} maintains
overall performance comparable to the original
DSA on this challenging benchmark.

\paragraph{RULER.}
We evaluate long-context retrieval and reasoning
performance at sequence lengths ranging from 16K to 128K.
\method{} achieves average scores of 95.28 and 95.35
on DeepSeek-V3.2 and GLM-5.1 respectively, 
closely matching the original DSA scores of
95.37 and 95.45.
As the context length increases, performance
differences among indexing acceleration methods
become increasingly pronounced.
At 128K, \method{} achieves scores of 93.34 and
94.49 on the two models, only 0.35 and 0.48
percentage points below the original DSA, respectively.
In contrast, at 128K, all four competing methods exhibit performance drops exceeding one percentage point relative to DSA on both models.
Based on RULER scores averaged across the four evaluated context lengths, \method{}
exhibits the smallest performance gap relative to
the original DSA among the evaluated acceleration methods.
Notably, it is the only method that remains within
one percentage point of the original DSA on both models at each of the four evaluated context lengths.
These results demonstrate that our method
maintains task performance even at long
context lengths.

\subsubsection{Efficiency Results}
\label{sec:efficiency_results}
\begin{figure*}[t]
    \centering
    \includegraphics[width=\textwidth]{
        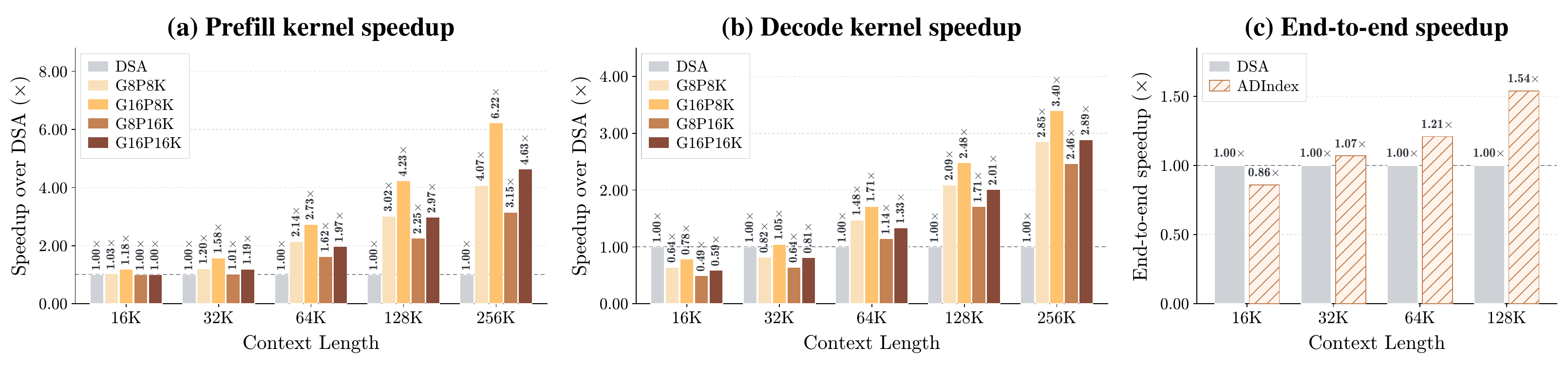
    }
    \caption{
        \textbf{Inference efficiency across different context lengths.}
        We report kernel speedup for prefill and decoding,
        together with end-to-end inference speedup.
        All speedups are measured relative to the original DSA baseline.
    }
    \label{fig:efficiency_speedup}
\end{figure*}

Figure~\ref{fig:efficiency_speedup} reports the speedups of
\method{} over the original DSA baseline across different
context lengths, covering both prefill and MTP
target-verification indexer kernels, as well as
end-to-end inference.
\suppressfloats[t]

\paragraph{Indexer Speedup.}
We benchmark our custom indexing kernels across different reuse configurations $(G,P)$ and
context lengths.
The DSA baseline consists primarily of full-context index
scoring and Top-$K$ selection, whereas \method{} comprises
shared token bank construction, query-wise refinement
within each group, and final Top-$K$ selection.
Figure~\ref{fig:efficiency_speedup} (a) and (b) report the 
indexing speedups relative to the original DSA baseline.
The gains become more pronounced at longer context lengths.
At 256K, \method{} achieves
$3.15\times$--$6.22\times$ speedups in prefill indexing and
$2.46\times$--$3.40\times$ speedups in decode indexing across
the four reuse configurations.
This improvement stems from reducing redundant full-context
scans: DSA performs full-context scoring independently for each
query, whereas \method{} constructs a shared token bank
for each group and limits query-wise refinement to this
substantially smaller token bank.
As the context grows, the cost of full-context scoring
increases, making our method
increasingly advantageous.
Stage-wise latency
breakdowns are provided in
Appendix~\ref{app:efficiency_speedup}.

\paragraph{End-to-End Speedup.}
Figure~\ref{fig:efficiency_speedup}(c) reports end-to-end
speedups relative to the original DSA baseline.
At shorter contexts, \method{} may be slower
than DSA. Specifically, when the context length is close
to the calibrated token bank budgets in certain layers,
shared token bank construction incurs additional overhead,
while query-specific refinement yields negligible
computational savings.
As the context grows, however, the benefits of reducing
redundant full-context scans become increasingly pronounced.
At a KV-cache length of 128K, \method{} achieves an
end-to-end speedup of $1.54\times$.
This trend is consistent with our complexity analysis:
longer contexts increase the cost of full-context indexing,
making shared token banks and query-specific refinement
increasingly effective at improving overall inference
efficiency.

\begingroup
\setlength{\columnsep}{10pt}
\setlength{\intextsep}{2pt}
\setlength{\aboverulesep}{1pt}
\setlength{\belowrulesep}{1.2pt}
\setlength{\cmidrulesep}{1.5pt}
\subsection{Ablation Studies}
\label{sec:ablation_studies}

We perform ablation studies to assess the effects
of layer-wise reuse calibration, group indexer training,
and the number of group indexer heads.
Additional ablations on deployment phase,
group representation construction, and the
layer-wise calibration criterion are provided
in Appendix~\ref{app:additional_ablations}.

\newcommand{\upgain}[1]{%
    \,{\fontsize{6}{7}\selectfont\textcolor{green!60!black}{$\uparrow$#1}}%
}
\newcommand{\downgain}[1]{%
    \,{\fontsize{6}{7}\selectfont\textcolor{red!70!black}{$\downarrow$#1}}%
}

\Needspace{16\baselineskip}
\par\medskip
\noindent\begin{minipage}{\linewidth}
    \captionsetup{type=table,position=top,justification=raggedright,singlelinecheck=false,skip=3pt}
    \caption{
        \textbf{Ablation on reuse configurations.}
        We report RULER scores (\%) and end-to-end speedups
        relative to the original DSA baseline on DeepSeek-V3.2
        at each context length.
        The best result in each column is highlighted in bold.
    }
    \label{tab:ablation_reuse_config}
    \centering
    \small
    \renewcommand{\textbf}[1]{{\bfseries\sffamily #1}}
    \setlength{\tabcolsep}{4pt}
    \renewcommand{\arraystretch}{1.05}

    \scalebox{0.79025}{%
    \begin{tabular*}{\dimexpr\linewidth*4000/3161\relax}{@{\extracolsep{\fill}}l*{4}{c}|*{4}{c}@{}}
        \toprule
        & \multicolumn{4}{c}{\textbf{RULER (\%)}}
        & \multicolumn{4}{c}{\textbf{End-to-End Speedup} ($\times$)} \\
        \cmidrule(lr){2-5}
        \cmidrule(l){6-9}
        \textbf{Setting}
        & \textbf{16K} & \textbf{32K} & \textbf{64K} & \textbf{128K}
        & \textbf{16K} & \textbf{32K} & \textbf{64K} & \textbf{128K} \\
        \midrule
        $(4,8\mathrm{K})$
        & 96.57 & \textbf{95.93} & \textbf{95.21} & \textbf{93.48}
        & \textbf{0.89} & 0.91 & 0.99 & 1.11 \\
        $(8,16\mathrm{K})$
        & 96.64 & 95.77 & 95.14 & 93.37
        & 0.69 & 1.03 & 1.13 & 1.30 \\
        Layer-wise
        & \textbf{96.81} & 95.81 & 95.16 & 93.34
        & 0.86 & \textbf{1.07} & \textbf{1.21} & \textbf{1.54} \\
        \bottomrule
    \end{tabular*}%
    }
\end{minipage}
\par\medskip

\paragraph{[Abl.1] Layer-wise reuse calibration improves inference efficiency.}
We first examine the benefits of selecting reuse
configurations independently for each layer.
To construct uniform baselines, we follow the
calibration procedure in Section~\ref{sec:reuse_setting},
with the additional constraint that all layers
share the same reuse configuration.
This yields $(G,P)=(4,8\mathrm{K})$ and
$(8,16\mathrm{K})$ for both models.
Uniform configurations must satisfy the capacity
constraints of all layers, potentially limiting
indexing efficiency.
In contrast, layer-wise calibration maximizes
each layer's indexing speedup
under its own capacity constraint.
As shown in Table~\ref{tab:ablation_reuse_config},
the layer-wise configuration remains within
0.24 percentage points of both uniform configurations
on RULER at every evaluated context length on DeepSeek-V3.2.
It also achieves the highest end-to-end speedup
among the three configurations at 32K, 64K, and 128K.
At 128K, it reaches $1.54\times$ relative to DSA,
compared with $1.11\times$ and $1.30\times$
for the two uniform configurations.
At 16K, it is slower than DSA ($0.86\times$) and the $(4,8\mathrm{K})$ configuration ($0.89\times$),
while providing speedups at 32K and longer contexts.
These results show that layer-wise calibration
improves inference efficiency over uniform configurations at longer contexts
while maintaining comparable task performance.

\Needspace{15\baselineskip}
\begin{wraptable}{r}{.57\textwidth}
    \captionsetup{position=top,justification=raggedright,singlelinecheck=false,skip=3pt}
\caption{
        \textbf{Ablation on group indexer training (RULER, \%).}
        Arrows indicate percentage-point changes
        relative to the variant without training.
    }
    \label{tab:ablation_learning}
\centering
    \small
    \renewcommand{\textbf}[1]{{\bfseries\sffamily #1}}
    \setlength{\tabcolsep}{2pt}
    \renewcommand{\arraystretch}{0.92}

    \scalebox{0.79025}{%
    \begin{tabular*}{\dimexpr\linewidth*4000/3161\relax}{@{\extracolsep{\fill}}l*{5}{c}@{}}
        \toprule
        & \multicolumn{4}{c}{\textbf{Context Length}}
        & \textbf{Overall} \\
        \cmidrule(lr){2-5}
        \cmidrule(l){6-6}

        \textbf{Training}
        & \textbf{16K}
        & \textbf{32K}
        & \textbf{64K}
        & \textbf{128K}
        & \textbf{AVG} \\
        \midrule

        \multicolumn{6}{c}{\textbf{DeepSeek-V3.2}} \\
        \midrule
        \textit{w/o training}
        & 96.41 & 95.16 & 94.63 & 91.66 & 94.47 \\

        \textit{w/ training}
        & 96.81\upgain{0.40}
        & 95.81\upgain{0.65}
        & 95.16\upgain{0.53}
        & 93.34\upgain{1.68}
        & 95.28\upgain{0.81} \\
        \midrule

        \multicolumn{6}{c}{\textbf{GLM-5.1}} \\
        \midrule
        \textit{w/o training}
        & 95.48 & 95.30 & 95.41 & 93.39 & 94.90 \\

        \textit{w/ training}
        & 95.84\upgain{0.36}
        & 95.57\upgain{0.27}
        & 95.51\upgain{0.10}
        & 94.49\upgain{1.10}
        & 95.35\upgain{0.45} \\

        \bottomrule
    \end{tabular*}%
    }
\end{wraptable}
\paragraph{[Abl.2] A distilled group indexer improves long-context performance.}
We compare two variants of \method{}:
\textit{w/o training}, which directly uses the frozen
original DSA indexer for shared token bank construction,
and \textit{w/ training}, which uses our lightweight
group indexer trained through group-level distillation.
As shown in Table~\ref{tab:ablation_learning},
the \textit{w/ training} variant improves the average RULER
score from 94.47 to 95.28 on DeepSeek-V3.2
and from 94.90 to 95.35 on GLM-5.1,
compared with the \textit{w/o training} variant.
The largest gains occur at 128K, reaching
1.68 and 1.10 percentage points, respectively.
These results demonstrate that the distilled group
indexer achieves better task performance than
directly using the frozen original DSA indexer,
while requiring only one-eighth as many indexing
heads for shared token bank construction. On DeepSeek-V3.2, the distilled group indexer
also achieves higher Top-2048 coverage than
the frozen mean-pooling baseline.
Further analysis is provided in
Appendix~\ref{app:shared_token_bank_coverage}.
\finishablationwrap

\Needspace{17\baselineskip}
\begin{wraptable}{r}{.50\textwidth}
    \captionsetup{position=top,justification=raggedright,singlelinecheck=false,skip=3pt}
\caption{
    \textbf{Ablation on the number of group indexer heads (RULER, \%).}
    The best score in each column
    is highlighted in bold.
}
\label{tab:ablation_heads}
\centering
    \small
    \renewcommand{\textbf}[1]{{\bfseries\sffamily #1}}
    \setlength{\tabcolsep}{2pt}
    \renewcommand{\arraystretch}{0.92}

    \scalebox{0.79025}{%
    \begin{tabular*}{\dimexpr\linewidth*4000/3161\relax}{@{\extracolsep{\fill}}l*{5}{c}@{}}
        \toprule
        & \multicolumn{4}{c}{\textbf{Context Length}}
        & \textbf{Overall} \\
        \cmidrule(lr){2-5}
        \cmidrule(l){6-6}

        \textbf{Num. Heads}
        & \textbf{16K}
        & \textbf{32K}
        & \textbf{64K}
        & \textbf{128K}
        & \textbf{Avg} \\
        \midrule

        \multicolumn{6}{c}{\textbf{DeepSeek-V3.2}} \\
        \midrule
        64 & 96.58 & 95.82 & 95.13 & 93.28 & 95.20 \\
        32 & 96.58 & \textbf{96.09} & \textbf{95.56} & 93.07 & \textbf{95.33} \\
        16 & 96.35 & 95.97 & 94.75 & 92.86 & 94.98 \\
        8  & \textbf{96.81} & 95.81 & 95.16 & \textbf{93.34} & 95.28 \\
        4  & 96.53 & 95.67 & 94.29 & 92.36 & 94.71 \\
        \midrule

        \multicolumn{6}{c}{\textbf{GLM-5.1}} \\
        \midrule
        32 & 95.87 & \textbf{95.73} & 95.42 & 94.59 & 95.40 \\
        16 & \textbf{95.97} & 95.20 & \textbf{95.82} & 94.52 & 95.38 \\
        8  & 95.94 & 95.72 & 95.63 & \textbf{94.63} & \textbf{95.48} \\
        4  & 95.84 & 95.57 & 95.51 & 94.49 & 95.35 \\

        \bottomrule
    \end{tabular*}%
    }
\end{wraptable}
\paragraph{[Abl.3] Fewer group indexer heads suffice.}
We finally investigate how the number of group indexer
heads affects task performance.
We vary the head count from 4 to 64 on DeepSeek-V3.2
and from 4 to 32 on GLM-5.1, while keeping the
original DSA indexer unchanged.
As shown in Table~\ref{tab:ablation_heads},
across the evaluated head counts, average performance
remains relatively stable: the highest and lowest
average scores differ by only 0.62 percentage points
on DeepSeek-V3.2 and 0.13 percentage points
on GLM-5.1.
To balance efficiency and task performance,
we adopt 8 and 4 group indexer heads for
DeepSeek-V3.2 and GLM-5.1, respectively.
Both configurations use one-eighth as many heads
as the corresponding original DSA indexer.
\finishablationwrap

\endgroup

\section{Conclusion}
\label{sec:conclusion}

We present \method{}, an adaptive and lightweight indexing method that reduces redundant full-prefix scans for token-level sparse attention through bank sharing across neighboring queries.
\method{} combines offline layer-wise calibration
of query group sizes and bank budgets
with a lightweight group indexer trained through
group-level distillation using fewer indexing heads.
During inference, \method{} uses shared token bank construction to
reduce redundant full-prefix scanning
and the original DSA indexer to perform
query-specific refinement to preserve fine-grained
token selection.
Evaluations on DeepSeek-V3.2 and GLM-5.1 show that
\method{} maintains near-lossless task performance across
LongBench, LongBench v2, and RULER.
Compared with standard DSA, \method{} delivers
a prefill kernel speedup of up to $6.22\times$ and a decoding kernel speedup of up to $3.40\times$ at 256K context among the layer-wise reuse configurations,
as well as an end-to-end serving speedup of $1.54\times$.

\newpage
{
\small
\bibliographystyle{plainnat}
\bibliography{references}
}

\clearpage
\beginappendix
\section{Analysis of Indexer Computation}
\label{app:indexer_complexity}

In this section, we analyze the indexer scoring operations
required by the original DSA and the token bank sharing mechanism of \method{}
during prefill and MTP target verification. We further quantify the indexing speedups under different query group sizes
and bank budgets.

\subsection{Cost Model and DSA Baseline}
\paragraph{Computational Cost Model.}
We adopt a simplified cost model in which computing the
indexer score between a single query and a single key
token is counted as one unit of computation.
This allows us to isolate the computational savings
from reducing the number of query--key comparisons.

Let $L$ denote the prefill sequence length, $G$ the
number of queries sharing a token bank, and $P$
the bank budget. We use $N$ to denote the
number of keys visible to a query, or to the anchor
query when analyzing a group. For MTP target verification,
each verification round processes a window of $m$ queries.
Let $R$ denote the number of verification rounds
and $N_j$ the number of keys visible to the first
query in round $j$. We assume $G$ divides $m$
and restrict grouping to individual verification rounds.
We compare the original DSA and our method under the
same verification workload, specified by $R$ and
$\{N_j\}_{j=1}^{R}$. During MTP verification,
short-context fallback is determined once per
verification round using the initial visible
prefix length $N_j$.

We account for scoring over all causally visible
post-anchor tokens. Our analysis focuses on indexer
scoring operations, excluding auxiliary costs such
as pooling, query projections, candidate selection,
and kernel launch overheads. In particular, we assign
the same unit cost to an original indexer score and a
group indexer score, irrespective of their respective
head counts. Thus, the resulting speedups characterize
reductions in scoring operations rather than
end-to-end wall-clock latency.

\paragraph{Baseline DSA Indexer.}
The original DSA indexer independently scores each
query against all causally visible keys. During prefill,
the number of available keys grows with the query
position. Approximating the discrete summation by an
integral, the total scoring cost is
\begin{equation}
\begin{aligned}
C_{\mathrm{base}}^{\mathrm{pre}}
&\approx \int_0^L N\,dN\\
&=\frac{L^2}{2}.
\end{aligned}
\label{eq:app_baseline_prefill}
\end{equation}

During MTP target verification, each round processes
$m$ queries in parallel. In round $j$, the $r$-th query,
$r\in\{0,\ldots,m-1\}$, has $N_j+r$ causally visible
keys. The total scoring cost over $R$ rounds is
\begin{equation}
\begin{aligned}
C_{\mathrm{base}}^{\mathrm{mtp}}
&=\sum_{j=1}^{R}\sum_{r=0}^{m-1}(N_j+r)\\
&=m\sum_{j=1}^{R}N_j+\frac{Rm(m-1)}{2}.
\end{aligned}
\label{eq:app_baseline_mtp}
\end{equation}

\subsection{Token Bank Sharing During Prefill and MTP Target Verification}
\paragraph{Bank Sharing During Prefill.}
Our method amortizes full-history indexer computation
across a group of $G$ neighboring queries. When the
anchor's visible context exceeds $P$, the group indexer
performs one full-prefix scan to construct a shared
token bank, after which the original DSA indexer
performs query-specific refinement for all $G$ queries.

For a group whose anchor has $N$ visible keys, bank
construction requires $N$ scoring operations, while
refinement requires $GP$ operations.
The $r$-th query in the group additionally scores
$r$ causally visible post-anchor tokens, where
$r\in\{0,\ldots,G-1\}$. Therefore, the group-level
scoring cost is
\begin{equation}
\begin{aligned}
C_{\mathrm{grp}}^{\mathrm{pre}}(N)
&=N+\sum_{r=0}^{G-1}(P+r)\\
&=N+GP+\frac{G(G-1)}{2}.
\end{aligned}
\label{eq:app_prefill_group}
\end{equation}
Amortizing this cost over the $G$ queries yields
\begin{equation}
c_{\mathrm{reuse}}^{\mathrm{pre}}(N)
=\frac{N}{G}+P+\frac{G-1}{2}.
\label{eq:app_prefill_per_query}
\end{equation}

When the anchor's visible context does not exceed
$P$, we fall back to the original DSA indexer.
For $L>P$, approximating the cumulative scoring
cost by continuous integration and neglecting
partial-group boundary effects gives
\begin{equation}
\begin{aligned}
C_{\mathrm{reuse}}^{\mathrm{pre}}
&\approx\int_0^P N\,dN\\
&\quad+\int_P^L
\left(\frac{N}{G}+P+\frac{G-1}{2}\right)dN\\
&=\frac{P^2}{2}+\frac{L^2-P^2}{2G}\\
&\quad+\left(P+\frac{G-1}{2}\right)(L-P).
\end{aligned}
\label{eq:app_reuse_prefill}
\end{equation}
For $L\le P$, the method falls back to the baseline,
so $C_{\mathrm{reuse}}^{\mathrm{pre}}=
C_{\mathrm{base}}^{\mathrm{pre}}$.

\paragraph{Bank Sharing During MTP Target Verification.}
The same bank construction and per-query refinement
are applied within each MTP verification round.
We partition the $m$ queries into $B=m/G$ disjoint
groups, each with its own candidate bank; no bank
is shared across verification rounds.
In our experiments, $m=16$, so
$G\in\{4,8,16\}$ gives four, two, or one
group per verification round, respectively.

Unlike prefill, the short-context fallback is
determined at the verification-window level.
Specifically, when the initial visible prefix length
$N_j$ does not exceed $P$, all $m$ queries in
round $j$ are processed by the original DSA
indexer. Otherwise, bank sharing is enabled for
all groups within the verification round.

When reuse is enabled, the anchor of group $b$,
where $b\in\{0,\ldots,B-1\}$, has $N_j+bG$
visible keys. Each group independently constructs
a token bank by scanning the full history
visible to its anchor, including tokens from
preceding groups within the same verification
round. All queries within the group, including
the anchor, then perform refinement,
with post-anchor tokens scored separately.

The scoring cost of an active verification round is
\begin{equation}
\begin{aligned}
C_{\mathrm{active},j}^{\mathrm{mtp}}
&=\sum_{b=0}^{B-1}
\left(
N_j+bG+GP+\frac{G(G-1)}{2}
\right)\\
&=\frac{m}{G}N_j+mP\\
&\quad+\frac{m}{2}
\left(\frac{m}{G}+G-2\right).
\end{aligned}
\label{eq:app_reuse_mtp_closed}
\end{equation}

Accounting for window-level short-context fallback,
we define the scoring cost of round $j$ as
\begin{equation}
\mathcal{C}_{\mathrm{win}}(N_j;G,P)=
\begin{cases}
mN_j+\dfrac{m(m-1)}{2}, & N_j\le P,\\[5pt]
C_{\mathrm{active},j}^{\mathrm{mtp}}, & N_j>P.
\end{cases}
\label{eq:app_mtp_window_cost}
\end{equation}

The total scoring cost over $R$ verification
rounds is therefore
\begin{equation}
C_{\mathrm{reuse}}^{\mathrm{mtp}}
=\sum_{j=1}^{R}
\mathcal{C}_{\mathrm{win}}(N_j;G,P).
\label{eq:app_reuse_mtp}
\end{equation}

This formulation accounts for both the
window-level fallback decision and the
independent full-prefix scans performed
by different query groups within each
active verification round.

\subsection{Indexing Speedups}
We define the indexing speedup as the
ratio of baseline to reuse scoring operations.
For prefill and MTP target verification, the
respective speedups are
\begin{equation}
\begin{aligned}
S_{\mathrm{pre}}
&=\frac{C_{\mathrm{base}}^{\mathrm{pre}}}
{C_{\mathrm{reuse}}^{\mathrm{pre}}},\\
S_{\mathrm{mtp}}
&=\frac{C_{\mathrm{base}}^{\mathrm{mtp}}}
{C_{\mathrm{reuse}}^{\mathrm{mtp}}}.
\end{aligned}
\label{eq:app_phase_speedups}
\end{equation}

We further define the joint  speedup
over the complete prefill and MTP target
verification workload as
\begin{equation}
S_{\mathrm{joint}}
=
\frac{
C_{\mathrm{base}}^{\mathrm{pre}}
+C_{\mathrm{base}}^{\mathrm{mtp}}
}{
C_{\mathrm{reuse}}^{\mathrm{pre}}
+C_{\mathrm{reuse}}^{\mathrm{mtp}}
}.
\label{eq:app_joint_speedup}
\end{equation}

Unlike standard autoregressive decoding, MTP
verifies multiple candidate tokens per round
but may accept only a subset. Consequently,
a fixed generation length does not uniquely
determine the number of verification rounds.
We therefore compare the original DSA and
our method under the same verification workload,
specified by $R$ and $\{N_j\}_{j=1}^{R}$.

For an active group with anchor context $N$,
subtracting the reuse cost from the baseline
cost yields
\begin{equation}
C_{\mathrm{base,grp}}-C_{\mathrm{reuse,grp}}
=(G-1)N-GP.
\label{eq:app_group_savings}
\end{equation}
Hence, reuse reduces the scoring count when
$N>GP/(G-1)$, a slightly stronger condition
than the activation threshold $N>P$.
As the context length grows with fixed $G$
and $P$, the per-group speedup approaches $G$.

\paragraph{Extension to Layer-wise Configurations.}
Different sparse-attention layers may adopt
different reuse configurations.
Let $\mathcal{L}$ denote the set of
sparse-attention layers, and let $(G_\ell,P_\ell)$
be the configuration assigned to layer $\ell$.

We denote the reuse scoring cost of layer $\ell$
during prefill or MTP target verification by
$C_{\mathrm{reuse},\ell}^{x}$, where
$x\in\{\mathrm{pre},\mathrm{mtp}\}$.
The joint cost for each layer is defined as
\begin{equation}
C_{\mathrm{reuse},\ell}^{\mathrm{joint}}
=
C_{\mathrm{reuse},\ell}^{\mathrm{pre}}
+
C_{\mathrm{reuse},\ell}^{\mathrm{mtp}}.
\label{eq:app_layerwise_joint_cost}
\end{equation}
Similarly, the baseline joint cost is
$C_{\mathrm{base}}^{\mathrm{joint}}
=C_{\mathrm{base}}^{\mathrm{pre}}
+C_{\mathrm{base}}^{\mathrm{mtp}}$.

We aggregate scoring operations across layers
before computing the  speedups:
\begin{equation}
\begin{aligned}
S_{\mathrm{layerwise}}^{x}
&=
\frac{
|\mathcal{L}|C_{\mathrm{base}}^{x}
}{
\displaystyle\sum_{\ell\in\mathcal{L}}
C_{\mathrm{reuse},\ell}^{x}
},\\
&\quad x\in\{\mathrm{pre},\mathrm{mtp},
\mathrm{joint}\}.
\end{aligned}
\label{eq:app_layerwise_speedup}
\end{equation}

These expressions account for the heterogeneous
group sizes and bank budgets selected
by layer-wise calibration, including
short-context fallback. The aggregate speedup
is computed from total scoring operations,
rather than by averaging per-layer speedups.

\subsection{Comparison Across Reuse Configurations}
\paragraph{Speedup Comparison.}
We evaluate indexer scoring speedups
under different reuse configurations for prefill,
MTP target verification, and their combined
inference workload.

We consider group sizes $G\in\{4,8,16\}$
and bank budgets
$P\in\{4,8,16\}\mathrm{K}$, with context lengths
of $16$, $32$, $64$, and $128\mathrm{K}$.
All lengths use binary units, i.e.,
$1\mathrm{K}=1024$ tokens.

For prefill, we use the continuous approximation
in Eq.~\eqref{eq:app_reuse_prefill}.
For MTP target verification, we evaluate a
single verification round ($R=1$), with
$m=16$ queries and the column context length
specifying the initial visible prefix length $N_1$.

For the joint analysis, we consider a controlled
verification workload with $R=32$ rounds and
$m=16$ queries per round.
We assume that each round produces exactly
$a=8$ tokens, including the final round,
resulting in a total generation length of
$D=Ra=256$ tokens.
The initial visible context length in round
$j$ is therefore given by
\begin{equation}
N_j=L+(j-1)a,\quad j=1,\ldots,R.
\label{eq:app_mtp_context_progression}
\end{equation}

Under this setting, the accumulated visible
history lengths satisfy
\begin{equation}
\sum_{j=1}^{R}N_j
=RL+\frac{aR(R-1)}{2}.
\label{eq:app_mtp_context_sum}
\end{equation}

The baseline and our method follow the same
verification workload. This setting provides
a controlled comparison rather
than assuming a measured MTP acceptance rate.

The calibration objective $S(G,P;L)$ in
Section~\ref{sec:reuse_setting} is defined as
$S_{\mathrm{joint}}$ evaluated under this fixed
verification workload, with $R=32$, $m=16$,
and $a=8$.

Table~\ref{tab:app_all_theoretical_speedups}
reports the corresponding speedups in three subpanels.

\begin{table*}[t]
\centering
\caption{Indexer scoring speedups ($\times$) across reuse configurations:
(a) prefill, (b) one MTP target verification round ($R=1$, $m=16$), and
(c) the joint prefill and MTP workload ($R=32$, $m=16$, $a=8$, $D=256$).
$G$ is the query group size and $P$ is the token bank budget; $1\mathrm{K}=1024$ tokens.
Column headings give the prefill context length in (a,c) and the keys visible
to the first query of the MTP round in (b).
All calculations include post-anchor scoring and short-context fallback,
with the same verification workload for the baseline and our method.}
\label{tab:app_all_theoretical_speedups}
\captionsetup[subtable]{format=plain,labelformat=parens,labelsep=space,
  font=small,labelfont=bf,justification=centering,singlelinecheck=true,skip=4pt}
\begin{subtable}[t]{0.32\textwidth}
\centering
\caption{Prefill}
\label{tab:app_theoretical_speedup}
\footnotesize
\setlength{\tabcolsep}{1.5pt}
\renewcommand{\arraystretch}{1.08}
\begin{tabular*}{\linewidth}{@{\extracolsep{\fill}}ccrrrr@{}}
\toprule
\multirow{2}{*}{$G$} & \multirow{2}{*}{$P$}
& \multicolumn{4}{c}{Context Length} \\
\cmidrule(lr){3-6}
& & 16K & 32K & 64K & 128K \\
\midrule
4 & 4K & 1.49 & 2.08 & 2.70 & 3.21 \\
4 & 8K & 1.07 & 1.49 & 2.08 & 2.70 \\
4 & 16K & 1.00 & 1.07 & 1.49 & 2.08 \\
\midrule
8 & 4K & 1.80 & 2.80 & 4.07 & 5.36 \\
8 & 8K & 1.18 & 1.80 & 2.80 & 4.07 \\
8 & 16K & 1.00 & 1.19 & 1.80 & 2.80 \\
\midrule
16 & 4K & 2.01 & 3.37 & 5.45 & 8.06 \\
16 & 8K & 1.25 & 2.01 & 3.38 & 5.45 \\
16 & 16K & 1.00 & 1.25 & 2.02 & 3.38 \\
\bottomrule
\end{tabular*}
\end{subtable}\hfill
\begin{subtable}[t]{0.32\textwidth}
\centering
\caption{MTP verification}
\label{tab:app_mtp_theoretical_speedup}
\footnotesize
\setlength{\tabcolsep}{1.5pt}
\renewcommand{\arraystretch}{1.08}
\begin{tabular*}{\linewidth}{@{\extracolsep{\fill}}ccrrrr@{}}
\toprule
\multirow{2}{*}{$G$} & \multirow{2}{*}{$P$}
& \multicolumn{4}{c}{Context Length} \\
\cmidrule(lr){3-6}
& & 16K & 32K & 64K & 128K \\
\midrule
4 & 4K & 2.00 & 2.67 & 3.20 & 3.56 \\
4 & 8K & 1.33 & 2.00 & 2.67 & 3.20 \\
4 & 16K & 1.00 & 1.33 & 2.00 & 2.67 \\
\midrule
8 & 4K & 2.67 & 4.00 & 5.33 & 6.40 \\
8 & 8K & 1.60 & 2.67 & 4.00 & 5.33 \\
8 & 16K & 1.00 & 1.60 & 2.67 & 4.00 \\
\midrule
16 & 4K & 3.20 & 5.33 & 7.99 & 10.66 \\
16 & 8K & 1.78 & 3.20 & 5.33 & 8.00 \\
16 & 16K & 1.00 & 1.78 & 3.20 & 5.33 \\
\bottomrule
\end{tabular*}
\end{subtable}\hfill
\begin{subtable}[t]{0.32\textwidth}
\centering
\caption{Prefill + MTP}
\label{tab:app_joint_theoretical_speedup}
\footnotesize
\setlength{\tabcolsep}{1.5pt}
\renewcommand{\arraystretch}{1.08}
\begin{tabular*}{\linewidth}{@{\extracolsep{\fill}}ccrrrr@{}}
\toprule
\multirow{2}{*}{$G$} & \multirow{2}{*}{$P$}
& \multicolumn{4}{c}{Context Length} \\
\cmidrule(lr){3-6}
& & 16K & 32K & 64K & 128K \\
\midrule
4 & 4K & 1.51 & 2.10 & 2.71 & 3.21 \\
4 & 8K & 1.08 & 1.50 & 2.09 & 2.70 \\
4 & 16K & 0.99 & 1.07 & 1.49 & 2.08 \\
\midrule
8 & 4K & 1.84 & 2.82 & 4.08 & 5.37 \\
8 & 8K & 1.20 & 1.82 & 2.81 & 4.08 \\
8 & 16K & 0.99 & 1.19 & 1.81 & 2.80 \\
\midrule
16 & 4K & 2.06 & 3.41 & 5.47 & 8.08 \\
16 & 8K & 1.28 & 2.04 & 3.40 & 5.46 \\
16 & 16K & 1.00 & 1.27 & 2.03 & 3.39 \\
\bottomrule
\end{tabular*}
\end{subtable}
\end{table*}

\paragraph{Analysis of Indexing Speedups.}
As shown in
Table~\ref{tab:app_all_theoretical_speedups}(\subref{tab:app_theoretical_speedup},\subref{tab:app_mtp_theoretical_speedup}),
the indexing speedup generally
increases with context length.
As the visible history grows, amortizing
full-prefix scanning across neighboring queries
becomes increasingly beneficial.

Larger query groups reduce the number of
full-prefix scans, while smaller bank budgets decrease the cost of query-specific
refinement. However, these configurations
must satisfy the estimated layer-specific
bank-capacity constraints.
Near the bank budget, the additional
bank-construction overhead may outweigh the
savings from reuse. Under window-level fallback,
MTP verification rounds with $N_j\le P$ retain
the original DSA scoring cost, while rounds
that have just crossed the activation threshold
may experience a slowdown.
These observations further motivate our
layer-wise reuse calibration.

Table~\ref{tab:app_joint_theoretical_speedup}
further reports the combined
speedups over prefill and MTP verification.
For $G=16$ and $P=4\mathrm{K}$, the joint
speedup increases from $2.06\times$ at 16K
context to $8.08\times$ at 128K context.

The joint speedups remain close to the
corresponding prefill speedups because
prefill dominates the cumulative indexer
scoring operations under the evaluated
workload. Nevertheless, the MTP contribution
is explicitly accounted for through the
fixed verification trace.

Overall, these results quantify how query group
size and bank budget affect indexer scoring cost.
Layer-wise reuse calibration selects efficient
reuse configurations subject to the estimated
bank-capacity constraints of individual attention layers.

Finally, we emphasize that these results
quantify reductions in indexer
scoring operations rather than observed latency.
Actual inference speedups also depend on the
group indexer architecture, memory access
patterns, kernel efficiency, and other
implementation-specific overheads.

\section{Additional Experiments and Analyses}
\label{app:additional_ablations}

We complement the ablations in
Section~\ref{sec:ablation_studies} with additional
experiments on deployment phase, group
representation construction, and the layer-wise
calibration criterion.
We conduct experiments on DeepSeek-V3.2 and GLM-5.1. Unless otherwise specified, we follow
the experimental settings described in
Section~\ref{sec:experimental_settings},
varying one component at a time while keeping
the remaining training and inference settings fixed.

\subsection{Deployment Phase}
\label{app:ablation_phase}

We investigate how deploying \method{} at
different inference stages affects task performance.
We consider three configurations:
applying \method{} only during prefill,
only during MTP target verification,
or during both stages.
The original DSA indexer is retained in stages
where \method{} is disabled.
Table~\ref{tab:ablation_phase} also includes
the original DSA as a reference.
On DeepSeek-V3.2, joint deployment achieves
an average RULER score of 95.28, compared
with 95.12 for prefill-only deployment and
95.16 for MTP-only deployment.
On GLM-5.1, MTP-only deployment achieves
the highest average score of 95.64,
while joint deployment obtains 95.35.
These results indicate that our method remains effective across different deployment phases.
We enable \method{} in both stages by default,
allowing token bank sharing during both prefill
and MTP target verification.
Under this setting, the average RULER scores
remain within 0.10 percentage points of
the original DSA on both models.
\begin{table}[!htb]
    \centering
    \small
    \setlength{\tabcolsep}{4pt}
    \renewcommand{\arraystretch}{1.15}

    \captionsetup{position=top}
    \caption{
        \textbf{Effect of the deployment phase on LongBench,
        LongBench v2, and RULER performance (\%).}
        MTP refers to MTP target verification.
        DSA uses the original indexer in both stages.
        Prefill only and MTP only apply our method exclusively
        during prefill and MTP target verification, respectively,
        while Both applies it in both stages.
        LongBench categories include Single-Document QA (SQA),
        Multi-Document QA (MQA), Summarization (Sum),
        Few-Shot Learning (FS), Synthetic (Syn), and
        Code Completion (Code).
        LongBench v2 results are reported for the Easy and Hard
        difficulty subsets and the Short, Medium, and Long
        context-length subsets.
        RULER results are reported at context lengths of
        16K, 32K, 64K, and 128K.
        Avg denotes the mean score across the six task categories
        for LongBench, the overall accuracy for LongBench v2,
        and the mean score across the four context lengths
        for RULER.
        Within each model, the best score in each column
        is highlighted in bold.
    }
    \label{tab:ablation_phase}

    \resizebox{\linewidth}{!}{%
        \begin{tabular}{l*{7}{c}|*{6}{c}|*{5}{c}}
            \toprule
            & \multicolumn{7}{c}{\textbf{LongBench}}
            & \multicolumn{6}{c}{\textbf{LongBench v2}}
            & \multicolumn{5}{c}{\textbf{RULER}} \\
            \cmidrule(lr){2-8}
            \cmidrule(lr){9-14}
            \cmidrule(lr){15-19}

            \textbf{Applied to}
            & \textbf{SQA}
            & \textbf{MQA}
            & \textbf{Sum}
            & \textbf{FS}
            & \textbf{Syn}
            & \textbf{Code}
            & \textbf{Avg}
            & \textbf{Easy}
            & \textbf{Hard}
            & \textbf{Short}
            & \textbf{Medium}
            & \textbf{Long}
            & \textbf{Avg}
            & \textbf{16K}
            & \textbf{32K}
            & \textbf{64K}
            & \textbf{128K}
            & \textbf{Avg} \\

            \midrule
            \multicolumn{19}{c}{\textbf{DeepSeek-V3.2}} \\
            \midrule

            DSA
            & 51.38 & 53.89 & 22.48 & 67.95 & 67.33
            & \textbf{73.40} & 56.07
            & 55.21 & 49.84 & 55.56 & 47.91 & 53.70 & 51.89
            & 96.72 & 95.78 & \textbf{95.28}
            & \textbf{93.69} & \textbf{95.37} \\

            Prefill only
            & \textbf{52.08} & 53.80 & 22.55 & 68.04
            & \textbf{67.69} & 72.96 & 56.19
            & 58.85 & \textbf{50.16} & 55.56
            & \textbf{49.77} & \textbf{57.41} & \textbf{53.48}
            & 96.40 & 95.54 & 95.18 & 93.35 & 95.12 \\

            MTP only
            & 51.39 & 53.98 & \textbf{22.68}
            & 68.08 & 67.33 & 72.31 & 55.96
            & 56.77 & 49.20 & 54.44 & 49.30 & 53.70 & 52.09
            & 96.58 & 95.76 & 94.72 & 93.58 & 95.16 \\

            Both
            & 51.84 & \textbf{54.33} & 22.54
            & \textbf{68.24} & 67.50 & 72.90 & \textbf{56.23}
            & \textbf{59.38} & 48.55 & \textbf{58.89}
            & 46.51 & 54.63 & 52.68
            & \textbf{96.81} & \textbf{95.81}
            & 95.16 & 93.34 & 95.28 \\

            \midrule
            \multicolumn{19}{c}{\textbf{GLM-5.1}} \\
            \midrule

            DSA
            & 50.22 & 55.74 & \textbf{23.95}
            & 68.66 & 67.10 & 70.92 & 56.10
            & 58.85 & 50.80 & 61.67 & 47.44 & 53.70 & 53.88
            & 95.66 & 95.61 & \textbf{95.54} & 94.97 & 95.45 \\

            Prefill only
            & \textbf{50.71} & \textbf{56.25} & 23.88
            & 68.15 & \textbf{67.50} & 69.78 & 56.04
            & \textbf{63.02} & 50.16 & 61.67
            & \textbf{49.77} & \textbf{54.63} & \textbf{55.07}
            & 95.38 & 95.42 & 94.18 & 94.33 & 94.83 \\

            MTP only
            & 49.86 & 55.71 & 23.80 & \textbf{68.83}
            & 67.46 & 69.48 & 55.86
            & 60.94 & 48.55 & 62.22 & 45.58 & 53.70 & 53.28
            & 95.66 & \textbf{96.78} & 95.01
            & \textbf{95.12} & \textbf{95.64} \\

            Both
            & 50.31 & 56.08 & 23.92 & 68.55 & 67.25
            & \textbf{70.94} & \textbf{56.18}
            & 57.29 & \textbf{51.45} & \textbf{64.44}
            & 46.05 & 50.93 & 53.68
            & \textbf{95.84} & 95.57 & 95.51 & 94.49 & 95.35 \\

            \bottomrule
        \end{tabular}%
    }
\end{table}

\subsection{Layer-wise Calibration Criterion}
\label{app:ablation_calibration_criterion}

We investigate the capacity criterion used
to determine the layer-specific reuse
configurations $(G_\ell,P_\ell)$.
For each layer and candidate group size,
we collect the cardinalities of the unions
of reference Top-$K$ indices across query
groups in the calibration set.
These reference indices are produced by
the original DSA indexer.
We compare three statistics for determining
configuration feasibility: the mean,
the 90th percentile ($P_{90}$), and the maximum.
A configuration is considered feasible
if its bank budget is no smaller than
the corresponding statistic.
For each criterion, we select the feasible
configuration that maximizes the 
indexing speedup.
The maximum-based criterion, adopted by default,
provides sufficient bank capacity for every
observed reference-index union in the
calibration set.
The mean- and percentile-based criteria relax
this capacity constraint and can admit more
aggressive reuse configurations, but the
resulting bank budgets may be insufficient
for some observed query groups.
As shown in
Table~\ref{tab:layer-wise_calibration_criterion},
the maximum-based criterion achieves the
highest average RULER score on both models.
Compared with the mean-based criterion,
it improves the average score by 0.55
and 0.44 percentage points on DeepSeek-V3.2
and GLM-5.1, respectively.
At 128K, the corresponding improvements
increase to 1.09 and 1.37 percentage points.
These results support the more conservative
capacity criterion used in our default
configuration.

\begin{table}[!htb]
    \centering
    \small
    \setlength{\tabcolsep}{4pt}
    \renewcommand{\arraystretch}{1.15}

    \captionsetup{position=top}
    \caption{
        \textbf{Effect of the layer-wise calibration criterion
        on LongBench, LongBench v2, and RULER performance (\%).}
        Mean, $P_{90}$, and Max use the mean,
        90th percentile, and maximum of the
        observed reference-index union sizes
        to determine configuration feasibility, respectively.
        LongBench categories include Single-Document QA (SQA),
        Multi-Document QA (MQA), Summarization (Sum),
        Few-Shot Learning (FS), Synthetic (Syn), and
        Code Completion (Code).
        LongBench v2 results are reported for the Easy and Hard
        difficulty subsets and the Short, Medium, and Long
        context-length subsets.
        RULER results are reported at context lengths of
        16K, 32K, 64K, and 128K.
        Avg denotes the mean score across the six task categories
        for LongBench, the overall accuracy for LongBench v2,
        and the mean score across the four context lengths
        for RULER.
        Within each model, the best score in each column
        is highlighted in bold, including ties.
    }
    \label{tab:layer-wise_calibration_criterion}

    \resizebox{\linewidth}{!}{%
        \begin{tabular}{l*{7}{c}|*{6}{c}|*{5}{c}}
            \toprule
            & \multicolumn{7}{c}{\textbf{LongBench}}
            & \multicolumn{6}{c}{\textbf{LongBench v2}}
            & \multicolumn{5}{c}{\textbf{RULER}} \\
            \cmidrule(lr){2-8}
            \cmidrule(lr){9-14}
            \cmidrule(lr){15-19}

            \textbf{Criterion}
            & \textbf{SQA}
            & \textbf{MQA}
            & \textbf{Sum}
            & \textbf{FS}
            & \textbf{Syn}
            & \textbf{Code}
            & \textbf{Avg}
            & \textbf{Easy}
            & \textbf{Hard}
            & \textbf{Short}
            & \textbf{Medium}
            & \textbf{Long}
            & \textbf{Avg}
            & \textbf{16K}
            & \textbf{32K}
            & \textbf{64K}
            & \textbf{128K}
            & \textbf{Avg} \\

            \midrule
            \multicolumn{19}{c}{\textbf{DeepSeek-V3.2}} \\
            \midrule

            Mean
            & 51.80 & 54.10 & \textbf{22.65}
            & \textbf{68.53} & 66.83 & 73.70 & 56.27
            & 57.81 & 47.59 & 55.56 & 48.37 & 50.93 & 51.49
            & 96.53 & 95.91 & 94.23 & 92.25 & 94.73 \\

            $P_{90}$
            & 51.81 & 53.82 & 22.55 & 68.20
            & \textbf{68.00} & \textbf{73.92} & \textbf{56.38}
            & 57.81 & \textbf{50.16} & 56.11
            & \textbf{51.16} & 51.85 & \textbf{53.08}
            & 96.44 & \textbf{95.98} & 94.71 & 93.07 & 95.05 \\

            Max
            & \textbf{51.84} & \textbf{54.33} & 22.54
            & 68.24 & 67.50 & 72.90 & 56.23
            & \textbf{59.38} & 48.55 & \textbf{58.89}
            & 46.51 & \textbf{54.63} & 52.68
            & \textbf{96.81} & 95.81 & \textbf{95.16}
            & \textbf{93.34} & \textbf{95.28} \\

            \midrule
            \multicolumn{19}{c}{\textbf{GLM-5.1}} \\
            \midrule

            Mean
            & 50.34 & \textbf{56.39} & 23.90
            & 68.47 & 67.18 & 70.16 & 56.07
            & 60.42 & 50.80 & 62.78
            & \textbf{49.30} & 50.93 & 54.47
            & 95.79 & 95.34 & 95.38 & 93.12 & 94.91 \\

            $P_{90}$
            & \textbf{50.56} & 55.93 & 23.79
            & 68.33 & \textbf{67.87} & 70.63 & \textbf{56.18}
            & \textbf{60.94} & 51.13 & \textbf{65.56}
            & 47.44 & \textbf{51.85} & \textbf{54.87}
            & \textbf{95.96} & \textbf{95.59}
            & 95.33 & 93.80 & 95.17 \\

            Max
            & 50.31 & 56.08 & \textbf{23.92}
            & \textbf{68.55} & 67.25
            & \textbf{70.94} & \textbf{56.18}
            & 57.29 & \textbf{51.45} & 64.44
            & 46.05 & 50.93 & 53.68
            & 95.84 & 95.57 & \textbf{95.51}
            & \textbf{94.49} & \textbf{95.35} \\

            \bottomrule
        \end{tabular}%
    }
\end{table}

\subsection{Group Representation Construction}
\label{app:ablation_group_representation}

We examine how different group representation
strategies affect RULER performance.
We compare representations constructed from
the first or last query in a group, mean
pooling over all queries, and the proposed
max--min--mean aggregation.
As shown in Table~\ref{tab:group_representation},
using only the first or last query leads to
substantial performance degradation on both models.
In contrast, mean pooling achieves average
scores of 94.95 and 94.92 on DeepSeek-V3.2
and GLM-5.1, respectively.
The max--min--mean aggregation further
improves these scores to 95.28 and 95.35,
yielding gains of 0.33 and 0.43 percentage
points, respectively.
It also achieves the highest score at every
evaluated context length on both models.
These results support aggregating information
across the query group and combining multiple
pooling statistics when constructing the
group representation.

\begin{table}[!htb]
\centering
\begin{minipage}{.70\textwidth}
    \captionsetup{position=top,justification=raggedright,singlelinecheck=false,skip=3pt}
\caption{
        \textbf{Effect of group representation
        construction on RULER performance (\%).}
        First and Last use the representation of
        the first and last query in each group,
        respectively.
        Avg denotes the mean score across the four
        context lengths.
        Within each model, the best score in each
        column is highlighted in bold.
    }
    \label{tab:group_representation}
\centering
    \small
    \setlength{\tabcolsep}{2pt}
    \renewcommand{\arraystretch}{1.05}

    \begin{tabular*}{\linewidth}{
        @{\extracolsep{\fill}}l*{5}{c}@{}
    }
        \toprule
        & \multicolumn{4}{c}{\textbf{Context Length}}
        & \textbf{Overall} \\
        \cmidrule(lr){2-5}
        \cmidrule(l){6-6}

        \textbf{Representation}
        & \textbf{16K}
        & \textbf{32K}
        & \textbf{64K}
        & \textbf{128K}
        & \textbf{Avg} \\
        \midrule

        \multicolumn{6}{c}{\textbf{DeepSeek-V3.2}} \\
        \midrule

        First
        & 92.23
        & 90.16
        & 89.58
        & 85.74
        & 89.43 \\

        Last
        & 91.94
        & 89.35
        & 90.12
        & 84.41
        & 88.96 \\

        Mean Pooling
        & 96.56
        & 95.47
        & 95.01
        & 92.77
        & 94.95 \\

        Max--Min--Mean
        & \textbf{96.81}
        & \textbf{95.81}
        & \textbf{95.16}
        & \textbf{93.34}
        & \textbf{95.28} \\
        \midrule

        \multicolumn{6}{c}{\textbf{GLM-5.1}} \\
        \midrule

        First
        & 91.37
        & 89.68
        & 87.29
        & 86.45
        & 88.70 \\

        Last
        & 92.01
        & 89.74
        & 88.13
        & 86.97
        & 89.21 \\

        Mean Pooling
        & 95.36
        & 95.24
        & 94.86
        & 94.23
        & 94.92 \\

        Max--Min--Mean
        & \textbf{95.84}
        & \textbf{95.57}
        & \textbf{95.51}
        & \textbf{94.49}
        & \textbf{95.35} \\

        \bottomrule
    \end{tabular*}
\end{minipage}
\end{table}

\subsection{Shared Token Bank Coverage}
\label{app:shared_token_bank_coverage}

We examine how well shared token bank construction
preserves the original DSA indexer's per-query
Top-2048 selections on DeepSeek-V3.2.
We compare three alternatives:
the aggregated teacher's Top-$K$ support
defined in Section~\ref{sec:training},
the distillation-driven group indexer,
and the frozen DSA indexer with mean pooling.
All 61 layers are evaluated under their calibrated
reuse settings.
Figure~\ref{fig:shared_token_bank_coverage}
reports coverage across query positions within each group.
For an overall summary, we average coverage
over query positions within each layer and then
average equally over all 61 layers.
These results indicate that the teacher support
retains most of the original DSA retrieval sets
and that the lightweight group indexer
preserves high coverage using only eight indexing heads.

\par\addvspace{6pt}
\noindent\begin{minipage}{\linewidth}
\captionsetup{type=figure}
    \centering
    \includegraphics[width=0.95\linewidth]
        {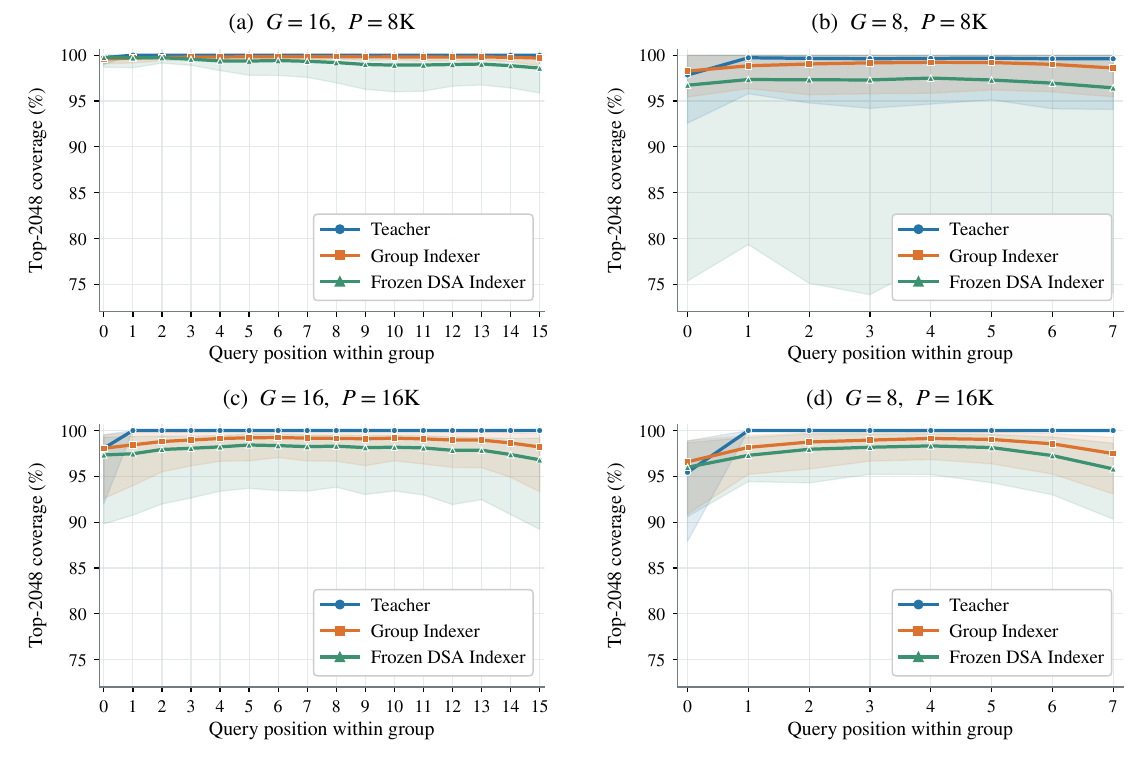}
    \caption{
        \textbf{Shared token bank coverage on DeepSeek-V3.2.}
        Teacher denotes the aggregated teacher's Top-$P$
        support; Group Indexer denotes the trained
        eight-head group indexer; Frozen DSA Indexer
        denotes the frozen indexer with mean pooling.
        (a)--(d) include 14, 15, 24, and 8 layers,
        respectively, according to their calibrated
        reuse settings.
        Shaded regions span the minimum and maximum
        of these layer-level values.
    }
    \label{fig:shared_token_bank_coverage}
\end{minipage}
\par\addvspace{6pt}

\subsection{Efficiency Speedup}
\label{app:efficiency_speedup}

Figures~\ref{fig:app_prefill_latency} and~\ref{fig:app_decode_latency}
complement Figure~\ref{fig:efficiency_speedup} with absolute single-layer
indexer latency. The latency advantage of \method{} generally grows with context
length, while bank construction overhead can offset the savings at short
contexts.
Figures~\ref{fig:app_prefill_proportion} and~\ref{fig:app_decode_proportion}
show the six-stage breakdown of \method{}, measured separately using
profiler kernel durations and normalized within each configuration.
Group scoring and token bank construction generally account for a larger
share at longer contexts. Refinement scoring includes candidate packing,
and final Top-$2048$ includes mapping and output handling.

\par\addvspace{6pt}
\noindent\begin{minipage}{\linewidth}
\captionsetup{type=figure}
    \centering
    \includegraphics[width=\linewidth,trim=0 5bp 0 7bp,clip]{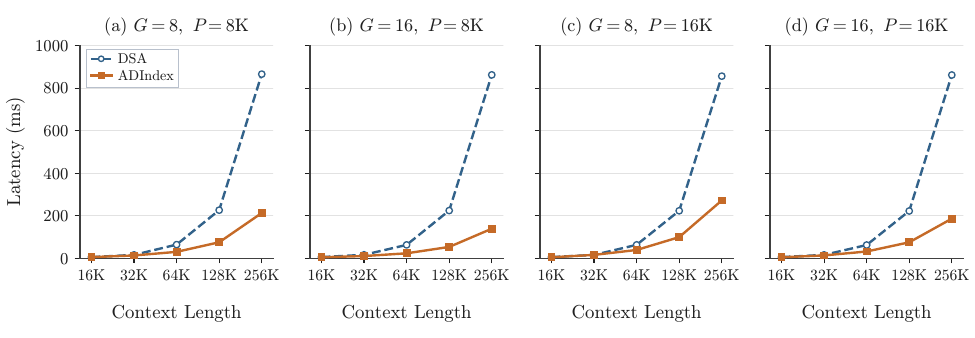}
    \caption{\textbf{Prefill indexer latency.}
        Each panel compares DSA and \method{} under a fixed reuse
        configuration $(G,P)$.}
    \label{fig:app_prefill_latency}
\end{minipage}
\par\addvspace{6pt}

\par\addvspace{6pt}
\noindent\begin{minipage}{\linewidth}
\captionsetup{type=figure}
    \centering
    \includegraphics[width=\linewidth,trim=0 5bp 0 7bp,clip]{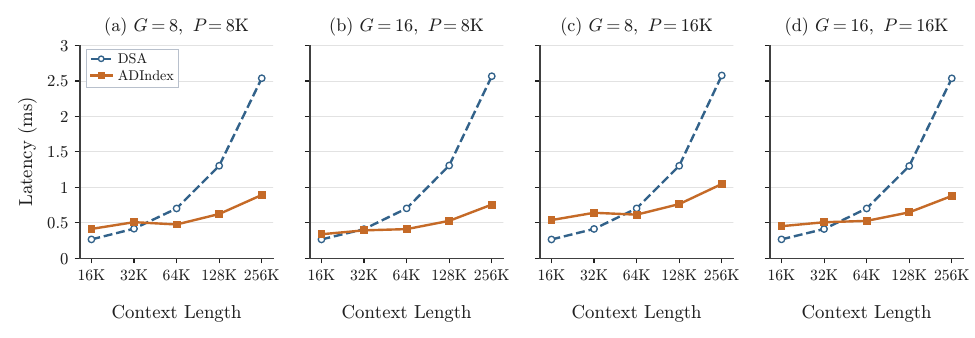}
    \caption{\textbf{Indexer latency during MTP target verification.}
        Each panel compares DSA and \method{} under a fixed reuse
        configuration $(G,P)$. }
    \label{fig:app_decode_latency}
\end{minipage}
\par\addvspace{6pt}

\par\addvspace{6pt}
\noindent\begin{minipage}{\linewidth}
\captionsetup{type=figure}
    \centering
    \includegraphics[width=\linewidth,trim=0 10bp 0 3bp,clip]{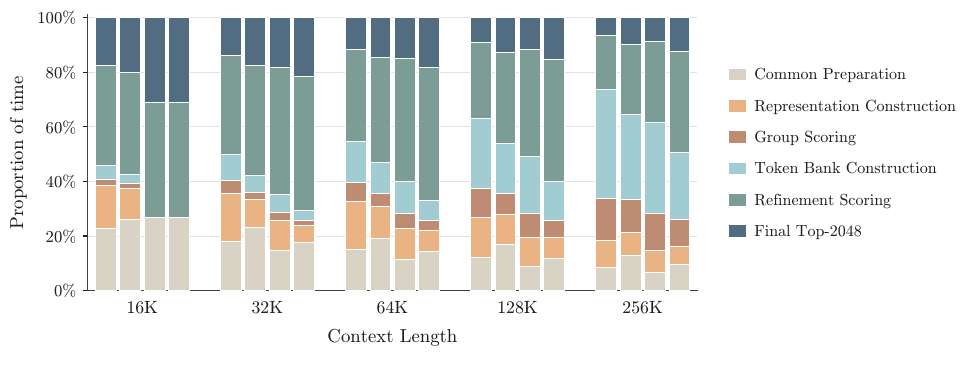}
    \caption{\textbf{Prefill stage breakdown of \method{}.}
        Within each context-length group, the four bars correspond,
        from left to right, to $(G,P)=(8,8\mathrm{K})$,
        $(16,8\mathrm{K})$, $(8,16\mathrm{K})$, and $(16,16\mathrm{K})$.
        Each bar sums to 100\% of its profiled kernel time.}
    \label{fig:app_prefill_proportion}
\end{minipage}
\par\addvspace{6pt}

\par\addvspace{6pt}
\noindent\begin{minipage}{\linewidth}
\captionsetup{type=figure}
    \centering
    \includegraphics[width=\linewidth,trim=0 10bp 0 3bp,clip]{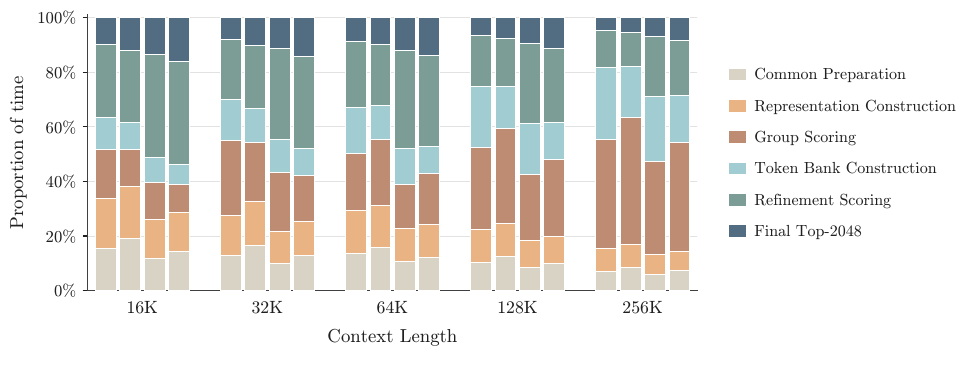}
    \caption{\textbf{Decode stage breakdown of \method{}.}
        Within each context-length group, the four bars correspond,
        from left to right, to $(G,P)=(8,8\mathrm{K})$,
        $(16,8\mathrm{K})$, $(8,16\mathrm{K})$, and $(16,16\mathrm{K})$.
        Each bar sums to 100\% of its profiled kernel time.}
    \label{fig:app_decode_proportion}
\end{minipage}
\par\addvspace{6pt}
\Needspace{6\baselineskip}
\section{Experimental and Implementation Details}

\providecommand{\appTBD}[1]{\textbf{[TBD: #1]}}

\subsection{Benchmark Details}
\label{app:benchmark details}

\paragraph{LongBench.}
LongBench~\citep{bai2024longbench} is a bilingual benchmark
for long-context understanding. The full benchmark contains
21 datasets spanning English, Chinese, and code, organized
into six categories: single-document question answering,
multi-document question answering, summarization, few-shot
learning, synthetic tasks, and code completion. We follow
the official task-specific evaluation protocols and report
the six category-level scores and their arithmetic mean.

\paragraph{LongBench v2.}
LongBench v2~\citep{bai2025longbenchv2} evaluates understanding
and reasoning over realistic long contexts through
503 multiple-choice questions. It covers single-document
question answering, multi-document question answering, long
in-context learning, long-dialogue history understanding,
code repository understanding, and long structured data
understanding. The original contexts range from approximately
8K to 2M words. We cap the input length at 128K tokens and
evaluate both models in non-thinking mode. We report overall
accuracy and the results on the Easy/Hard and
Short/Medium/Long subsets using the official evaluation
protocol.

\paragraph{RULER.}
RULER~\citep{hsieh2024ruler} evaluates long-context capabilities
with configurable sequence lengths and task complexity.
Its standard suite comprises 13 tasks covering retrieval,
multi-hop tracing, aggregation, and question answering.
These tasks assess both information retrieval and the
ability to combine information distributed throughout a
long context. We evaluate at 16K, 32K, 64K, and 128K tokens
under identical settings across compared methods, and report
the score at each length and their arithmetic mean.

\subsection{Training Data Details}
\label{app:training data details}

\paragraph{Source corpus and offline feature extraction.}
We use the publicly available LiteLong+NExtLong 128K
corpus,\footnote{\url{https://huggingface.co/datasets/caskcsg/Litelong_Nextlong_128k}}
which is constructed from FineWeb-Edu and Cosmopedia V2.
LiteLong retrieves topically related documents using BM25,
while NExtLong extends document chunks with hard negative
distractors to strengthen long-range
dependencies~\citep{jia2026litelong,gao2025nextlong}.
Their combination provides long-context sequences for
extracting the layer-specific representations used in
our training.

We randomly select 1,000 source sequences from this corpus
for offline hidden-state extraction.
For each frozen pretrained model, we cache the input
representations of all Transformer layers.
These cached sequences serve as the source of group-level
training instances for the group indexer.

The cached representations correspond to the residual stream
immediately before each layer's input RMSNorm.
When a cache is organized by layer outputs, the output of
layer $\ell-1$ is used as the input to layer $\ell$;
the first layer receives the token embeddings.
During training, we apply the frozen input RMSNorm and
query-latent projections to obtain the normalized hidden
representations and query latents required by the group indexer.
Historical indexing keys are computed using the frozen
indexer and cached for reuse.

\paragraph{Anchor-based subsampling.}
We construct the optimization dataset by forming candidate
groups of consecutive cached hidden states at a fixed stride
and randomly sampling from these groups.
Each hidden-state group corresponds to a group of queries
at the same token positions, and its starting position is
termed the \emph{query-group anchor}.
For a sequence of length $L$ and a layer-specific group
size $G_\ell$, the eligible anchors are
\begin{equation}
\mathcal{A}_{L,\ell}
=
\left\{
t=32768+16j
\;\middle|\;
j\in\mathbb{Z}_{\geq 0},
\quad t+G_\ell-1<L
\right\}.
\end{equation}
Here, token positions are zero-indexed.
Each sampled anchor $t$ specifies a training instance whose
group input consists of the cached hidden states at positions
$t,\ldots,t+G_\ell-1$.
We allocate per-sequence sampling quotas, sample anchors
using stratified random sampling over the eligible grids,
and shuffle the resulting instances.
For each layer, training uses a total of 256,000 sampled
group instances, corresponding to 1,000 optimizer updates
with a global batch size of 256.
Thus, the 1,000 source sequences define the corpus used for
feature extraction, while the optimization data consist of
sampled query groups rather than an exhaustive enumeration
of all token positions.
Subsampling applies only to the group anchors:
all historical indexing keys at positions $0,\ldots,t$
remain available, with their original position indices
preserved.

\paragraph{Calibration and validation data.}
For calibration, we randomly sample 100 source sequences
from the same corpus, excluding all source sequences used
for training.
This calibration set is used to determine the layer-specific
group sizes and bank budgets.

For each layer, the training and validation sets contain
256,000 and 64,000 sampled query groups, respectively,
yielding a training-to-validation ratio of $4{:}1$
at the group level.

\subsection{Group Indexer Training Details}
\label{app:group indexer training details}

\paragraph{Trainable modules and initialization.}
We train each layer-specific group indexer independently
while freezing the pretrained backbone and original
DSA indexer. The default configuration uses eight student
indexing heads for DeepSeek-V3.2 and four for GLM-5.1,
with a head dimension of 128.
We apply max--min--mean pooling to the normalized query
latents and mean pooling to the normalized hidden
representations. Only the pooling projection
$\mathbf{W}_{\mathrm{pool}}$, query projection $\mathbf{W}_Q^S$,
and gating projection $\mathbf{W}_G^S$ are optimized.

The query and gating projection matrices are bias-free
and initialized independently from
$\mathcal{U}(-1/\sqrt{d_{\mathrm{in}}},
1/\sqrt{d_{\mathrm{in}}})$, where $d_{\mathrm{in}}$
is the corresponding input dimension.
They directly parameterize the student rather than
residual updates to the original indexer weights.
The pooling projection is initialized as $[0,0,I]$
so that its initial output equals mean pooling.
Rotary position embeddings for the group-level query
are evaluated at the anchor position.
Following the main text, the group-level query and gate
are computed using Eq.~\eqref{eq:student_query_gate}. 
The ReLU is applied to each query--key dot product
before the weighted sum over heads.

\paragraph{Teacher construction and objective.}
We follow the teacher construction and distillation objective
in Section~\ref{sec:training}, with the teacher distribution
and loss defined in Eqs.~\eqref{eq:group_teacher_distribution}
and~\eqref{eq:group_distillation_loss}, respectively.
We set the teacher and student temperatures to $\tau_T=3$
and $\tau_S=1$.
For score standardization, the mean and standard deviation
are computed over the anchor-visible history, and the
standard deviation is clamped from below at $10^{-6}$
for numerical stability.
Teacher targets are computed from the cached inputs using
the frozen DSA indexer, and the per-group KL losses are
averaged within each mini-batch.
No forced sink-token reservation or additional local
window is used during training.
\paragraph{Optimization.}
We use AdamW with $\beta_1=0.9$, $\beta_2=0.95$,
zero weight decay, and a peak learning rate of
$2\times10^{-4}$.
Each layer is trained for 1,000 optimizer steps,
with 100 steps of linear warmup followed by cosine
decay toward 10\% of the peak learning rate over
a 1,000-step schedule.
Gradients are clipped to a global norm of 1.0.

Training uses eight data-parallel workers, each processing
32 anchors per update, giving a global batch size of 256
and 256,000 sampled anchor visits per layer.
Gradients are averaged across workers, and trainable
parameters and AdamW states are maintained in FP32.
Each worker maintains a cache of eight resident training
sequences and replaces sequences as their sampling
quotas are exhausted.

\paragraph{Training Convergence.}
We further examine the optimization behavior of
the group indexer with different numbers of
indexing heads.
In the plots, $H$ denotes the group indexer's head count $H_S^I$.
Figures~\ref{fig:loss_dsv32_l15}--\ref{fig:loss_glm51_l60}
present the training loss over 1,000 optimization
steps for representative layers of DeepSeek-V3.2
and GLM-5.1 under different reuse configurations.

Across the evaluated layers and head counts,
the training loss decreases substantially
during the initial optimization steps and
gradually stabilizes.
Although the convergence trajectories and
final loss values vary across configurations,
the group indexer exhibits stable optimization
behavior even with substantially fewer
indexing heads.

\par\begingroup
\setlength{\parskip}{0pt}
\par\addvspace{4pt}
\nointerlineskip
\noindent\begin{minipage}{\linewidth}
\captionsetup{type=figure}
    \centering
    \includegraphics[width=\textwidth,trim=0 9bp 0 6bp,clip]{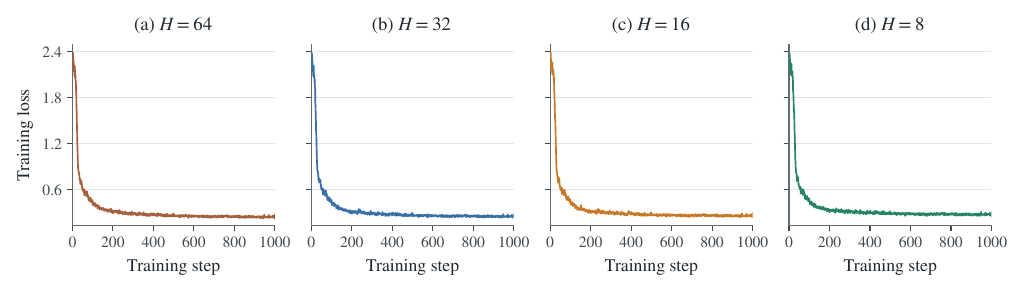}
    \caption{
        \textbf{Group indexer training loss on DeepSeek-V3.2 (Layer 15).}
        The reuse configuration is $(G,P)=(8,8\mathrm{K})$.
        Panels (a)--(d) correspond to
        $H_S^I\in\{64,32,16,8\}$ indexing heads, respectively.
    }
    \label{fig:loss_dsv32_l15}
\end{minipage}
\par\addvspace{4pt}

\par\addvspace{4pt}
\nointerlineskip
\noindent\begin{minipage}{\linewidth}
\captionsetup{type=figure}
    \centering
    \includegraphics[width=\textwidth,trim=0 9bp 0 6bp,clip]{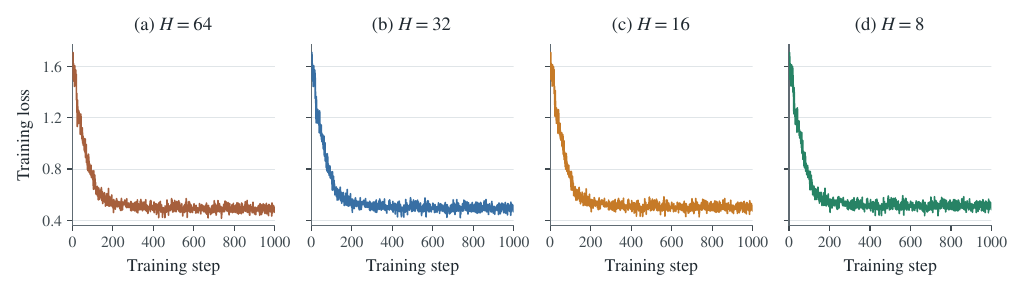}
    \caption{
        \textbf{Group indexer training loss on DeepSeek-V3.2 (Layer 17).}
        The reuse configuration is $(G,P)=(8,16\mathrm{K})$.
        Panels (a)--(d) correspond to
        $H_S^I\in\{64,32,16,8\}$ indexing heads, respectively.
    }
    \label{fig:loss_dsv32_l17}
\end{minipage}
\par\addvspace{4pt}

\par\addvspace{4pt}
\nointerlineskip
\noindent\begin{minipage}{\linewidth}
\captionsetup{type=figure}
    \centering
    \includegraphics[width=\textwidth,trim=0 9bp 0 6bp,clip]{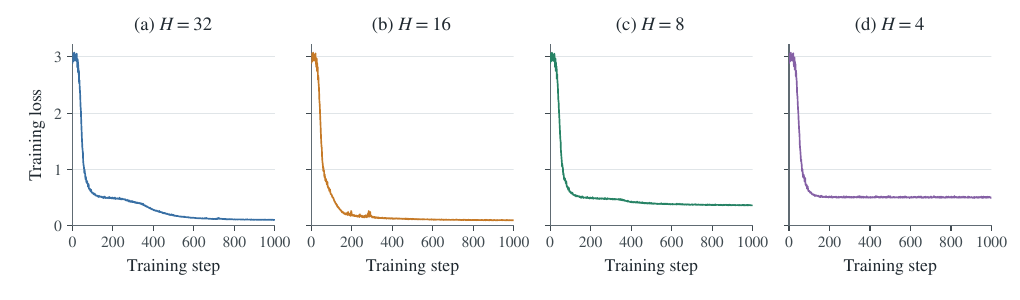}
    \caption{
        \textbf{Group indexer training loss on GLM-5.1 (Layer 2).}
        The reuse configuration is $(G,P)=(16,4\mathrm{K})$.
        Panels (a)--(d) correspond to
        $H_S^I\in\{32,16,8,4\}$ indexing heads, respectively.
    }
    \label{fig:loss_glm51_l2}
\end{minipage}
\par\addvspace{4pt}

\par\addvspace{4pt}
\nointerlineskip
\noindent\begin{minipage}{\linewidth}
\captionsetup{type=figure}
    \centering
    \includegraphics[width=\textwidth,trim=0 9bp 0 6bp,clip]{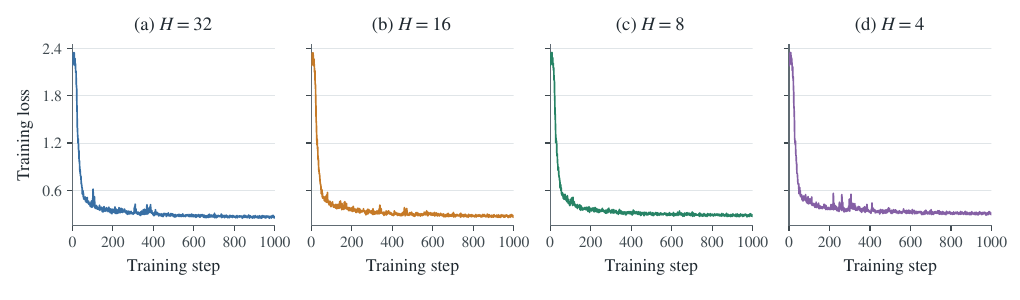}
    \caption{
        \textbf{Group indexer training loss on GLM-5.1 (Layer 59).}
        The reuse configuration is $(G,P)=(16,8\mathrm{K})$.
        Panels (a)--(d) correspond to
        $H_S^I\in\{32,16,8,4\}$ indexing heads, respectively.
    }
    \label{fig:loss_glm51_l59}
\end{minipage}
\par\addvspace{4pt}

\par\addvspace{4pt}
\nointerlineskip
\noindent\begin{minipage}{\linewidth}
\captionsetup{type=figure}
    \centering
    \includegraphics[width=\textwidth,trim=0 9bp 0 6bp,clip]{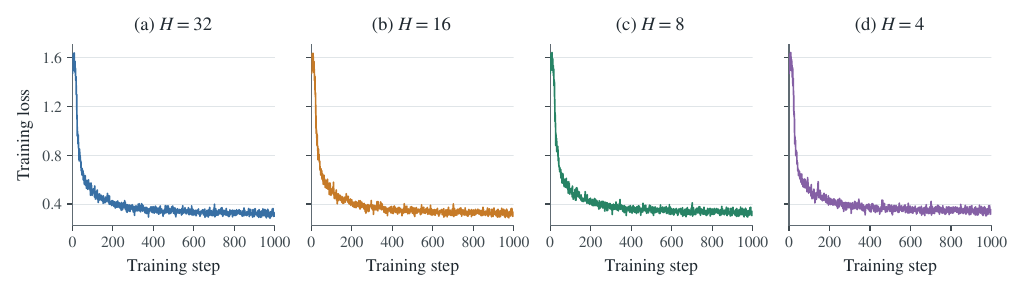}
    \caption{
        \textbf{Group indexer training loss on GLM-5.1 (Layer 60).}
        The reuse configuration is $(G,P)=(16,16\mathrm{K})$.
        Panels (a)--(d) correspond to
        $H_S^I\in\{32,16,8,4\}$ indexing heads, respectively.
    }
    \label{fig:loss_glm51_l60}
\end{minipage}
\par\addvspace{4pt}
\par\endgroup

\Needspace{10\baselineskip}
\subsection{Layer-wise Reuse Configurations}
\label{app:layer-wise reuse configurations}
\paragraph{Group configurations and available context.}
Each layer uses its calibrated
$G_\ell\in\{4,8,16\}$ and
$P_\ell\in\{4096,8192,16384\}$,
with a final per-query retrieval budget of $K=2048$.
The complete layer-wise configurations are provided in
Tables~\ref{tab:dsv32_prefill_max}--\ref{tab:glm51_prefill_mean}.

The inputs are the query representations jointly
processed during prefill or MTP target verification,
while token bank retrieval is restricted to
positions visible at the group anchor.
Positions that become visible after the anchor
are added during query-wise refinement.
\par\begingroup
\setlength{\parskip}{0pt}
\par\addvspace{4pt}
\nointerlineskip
\noindent\begin{minipage}{\linewidth}
\captionsetup{type=table}
\centering
\footnotesize
\setlength{\tabcolsep}{3pt}
\renewcommand{\arraystretch}{1.05}
\setlength{\abovecaptionskip}{0pt}
\setlength{\belowcaptionskip}{4pt}

\caption{Layer-wise reuse configurations for
DeepSeek-V3.2 calibrated using the maximum
DSA reference Top-2048 index union sizes.}
\label{tab:dsv32_prefill_max}
\begin{tabularx}{\linewidth}{
        @{}l c >{\raggedright\arraybackslash}X@{}}
        \toprule
        \textbf{Setting $(G,P)$}
        & \textbf{Count}
        & \textbf{Layers} \\
        \midrule

        $(16,8\mathrm{K})$
        & 14
        & 2, 3, 5--8, 10, 12--14, 18,
          22, 37, 44 \\

        $(8,8\mathrm{K})$
        & 15
        & 0, 1, 9, 15, 23, 25, 34, 36,
          41, 42, 47, 53, 54, 58, 60 \\

        $(16,16\mathrm{K})$
        & 24
        & 4, 11, 16, 19, 20, 24, 26--32,
          35, 38--40, 48, 50--52, 56,
          57, 59 \\

        $(8,16\mathrm{K})$
        & 8
        & 17, 21, 33, 43, 45, 46, 49, 55 \\

        \bottomrule
    \end{tabularx}
\end{minipage}
\par\addvspace{4pt}

\par\addvspace{4pt}
\nointerlineskip
\noindent\begin{minipage}{\linewidth}
\captionsetup{type=table}
\centering
\footnotesize
\setlength{\tabcolsep}{3pt}
\renewcommand{\arraystretch}{1.05}
\setlength{\abovecaptionskip}{0pt}
\setlength{\belowcaptionskip}{4pt}

\caption{Layer-wise reuse configurations for
DeepSeek-V3.2 calibrated using the 90th percentile of
DSA reference Top-2048 index union sizes.}
\label{tab:dsv32_prefill_p90}
\begin{tabularx}{\linewidth}{
        @{}l c >{\raggedright\arraybackslash}X@{}}
        \toprule
        \textbf{Setting $(G,P)$}
        & \textbf{Count}
        & \textbf{Layers} \\
        \midrule

        $(16,4\mathrm{K})$
        & 3
        & 2, 3, 5 \\

        $(16,8\mathrm{K})$
        & 24
        & 0, 1, 6--10, 12--15, 18, 22, 23,
          25, 34, 36, 37, 41, 42, 44, 48,
          58, 60 \\

        $(8,8\mathrm{K})$
        & 30
        & 4, 11, 16, 19--21, 24, 26--32,
          35, 38--40, 45, 47, 49--57, 59 \\

        $(16,16\mathrm{K})$
        & 2
        & 33, 43 \\

        $(8,16\mathrm{K})$
        & 2
        & 17, 46 \\

        \bottomrule
    \end{tabularx}
\end{minipage}
\par\addvspace{4pt}

\par\addvspace{4pt}
\nointerlineskip
\noindent\begin{minipage}{\linewidth}
\captionsetup{type=table}
\centering
\footnotesize
\setlength{\tabcolsep}{3pt}
\renewcommand{\arraystretch}{1.05}
\setlength{\abovecaptionskip}{0pt}
\setlength{\belowcaptionskip}{4pt}

\caption{Layer-wise reuse configurations for
DeepSeek-V3.2 calibrated using the mean of
DSA reference Top-2048 index union sizes.}
\label{tab:dsv32_prefill_mean}
\begin{tabularx}{\linewidth}{
        @{}l c >{\raggedright\arraybackslash}X@{}}
        \toprule
        \textbf{Setting $(G,P)$}
        & \textbf{Count}
        & \textbf{Layers} \\
        \midrule

        $(16,4\mathrm{K})$
        & 4
        & 2, 3, 5, 7 \\

        $(16,8\mathrm{K})$
        & 43
        & 0, 1, 4, 6, 8--10, 12--16, 18, 19,
          22, 23, 25, 28, 30--32, 34--38,
          40--42, 44, 45, 47, 48, 50--54,
          56--60 \\

        $(8,8\mathrm{K})$
        & 13
        & 11, 20, 21, 24, 26, 27, 29, 33,
          39, 43, 46, 49, 55 \\

        $(16,16\mathrm{K})$
        & 1
        & 17 \\

        \bottomrule
    \end{tabularx}
\end{minipage}
\par\addvspace{4pt}

\par\addvspace{4pt}
\nointerlineskip
\noindent\begin{minipage}{\linewidth}
\captionsetup{type=table}
\centering
\footnotesize
\setlength{\tabcolsep}{3pt}
\renewcommand{\arraystretch}{1.05}
\setlength{\abovecaptionskip}{0pt}
\setlength{\belowcaptionskip}{4pt}

\caption{Layer-wise reuse configurations for
GLM-5.1 calibrated using the maximum
DSA reference Top-2048 index union sizes.}
\label{tab:glm51_prefill_max}
\begin{tabularx}{\linewidth}{
        @{}l c >{\raggedright\arraybackslash}X@{}}
        \toprule
        \textbf{Setting $(G,P)$}
        & \textbf{Count}
        & \textbf{Layers} \\
        \midrule

        $(16,4\mathrm{K})$
        & 1
        & 2 \\

        $(16,8\mathrm{K})$
        & 16
        & 0, 1, 3, 5--9, 15--17, 26,
          32, 40, 48, 59 \\

        $(8,8\mathrm{K})$
        & 32
        & 4, 13, 14, 18, 19, 21--25,
          28--31, 35, 36, 49, 53, 55,
          58, 61, 66--75, 77 \\

        $(16,16\mathrm{K})$
        & 13
        & 10, 11, 20, 33, 34, 38,
          41, 46, 54, 56, 57, 60, 63 \\

        $(8,16\mathrm{K})$
        & 16
        & 12, 27, 37, 39, 42--45,
          47, 50--52, 62, 64, 65, 76 \\

        \bottomrule
    \end{tabularx}
\end{minipage}
\par\addvspace{4pt}

\par\addvspace{4pt}
\nointerlineskip
\noindent\begin{minipage}{\linewidth}
\captionsetup{type=table}
\centering
\footnotesize
\setlength{\tabcolsep}{3pt}
\renewcommand{\arraystretch}{1.05}
\setlength{\abovecaptionskip}{0pt}
\setlength{\belowcaptionskip}{4pt}

\caption{Layer-wise reuse configurations for
GLM-5.1 calibrated using the 90th percentile of
DSA reference Top-2048 index union sizes.}
\label{tab:glm51_prefill_p90}
\begin{tabularx}{\linewidth}{
        @{}l c >{\raggedright\arraybackslash}X@{}}
        \toprule
        \textbf{Setting $(G,P)$}
        & \textbf{Count}
        & \textbf{Layers} \\
        \midrule

        $(16,4\mathrm{K})$
        & 3
        & 0, 2, 5 \\

        $(16,8\mathrm{K})$
        & 38
        & 1, 3, 4, 6--9, 13--17, 19, 20,
          22, 26, 28, 29, 31, 32, 35, 36,
          40, 48, 49, 53--55, 57--59, 61,
          66--69, 71, 72 \\

        $(8,8\mathrm{K})$
        & 25
        & 10, 11, 18, 21, 23--25, 30, 34,
          38, 41--44, 46, 47, 51, 56,
          60, 70, 73--77 \\

        $(16,16\mathrm{K})$
        & 12
        & 12, 27, 33, 37, 39, 45, 50,
          52, 62--65 \\

        \bottomrule
    \end{tabularx}
\end{minipage}
\par\addvspace{4pt}

\par\addvspace{4pt}
\nointerlineskip
\noindent\begin{minipage}{\linewidth}
\captionsetup{type=table}
\centering
\footnotesize
\setlength{\tabcolsep}{3pt}
\renewcommand{\arraystretch}{1.05}
\setlength{\abovecaptionskip}{0pt}
\setlength{\belowcaptionskip}{4pt}

\caption{Layer-wise reuse configurations for
GLM-5.1 calibrated using the mean of
DSA reference Top-2048 index union sizes.}
\label{tab:glm51_prefill_mean}
\begin{tabularx}{\linewidth}{
        @{}l c >{\raggedright\arraybackslash}X@{}}
        \toprule
        \textbf{Setting $(G,P)$}
        & \textbf{Count}
        & \textbf{Layers} \\
        \midrule

        $(16,4\mathrm{K})$
        & 6
        & 0--3, 5, 8 \\

        $(16,8\mathrm{K})$
        & 47
        & 4, 6, 7, 9, 10, 13--22, 24--26,
          28--32, 35, 36, 40, 42, 48, 49,
          53--55, 57--59, 61, 66--75, 77 \\

        $(8,8\mathrm{K})$
        & 25
        & 11, 12, 23, 27, 33, 34, 37--39,
          41, 43--47, 50--52, 56, 60,
          62--65, 76 \\

        \bottomrule
    \end{tabularx}
\end{minipage}
\par\addvspace{4pt}
\par\endgroup

\Needspace{6\baselineskip}
\subsection{Baseline Implementations}
\label{app:baseline details}

We compare \method{} with the original DSA indexer and four
indexing acceleration methods.
Both models are evaluated in non-thinking mode with
identical evaluation settings across methods within
each benchmark. Inference uses SGLang on NVIDIA B300
GPUs. The settings below describe our reproductions.

\paragraph{HISA.}
  HISA~\citep{xu2026hisa} first selects candidate blocks
  using pooled indexing keys and then applies token-level
  refinement within the selected blocks.
  Our SGLang implementation uses 128-token blocks, a budget
  of up to 64 candidate blocks per query, and mean
  pooling of indexing keys.
  The first block and the block containing the current query
  are always included and count toward the 64-block budget.
  No additional initial-token or local-window retention is enabled.
  Forced block retention applies only to candidate generation.
  Candidate tokens are rescored using all indexing heads,
  and the final retrieval budget is 2048 tokens per query,
  subject to causal masking.
  When at most 64 blocks are visible, all visible blocks
  are considered; when at most 2048 tokens are visible,
  all visible tokens are retained.
  HISA is enabled in all 61 layers of DeepSeek-V3.2 and
  all 78 layers of GLM-5.1 during both prefill and decoding.

\paragraph{MISA.}
MISA~\citep{zhou2026misa} uses block-level statistics
to route each query to a subset of indexing heads
for token-level scoring.
We use MISA with 8 active heads and a router block size
of 1024 tokens for both DeepSeek-V3.2 and GLM-5.1. The hierarchical
candidate budget is not applicable. The final retrieval budget is 2048 per query.
The method is enabled in all target model layers (61 in DeepSeek-V3.2 and 78 in GLM-5.1) during prefill and decoding.

\paragraph{IndexCache.}
IndexCache~\citep{bai2026indexcache} computes retrieval
indices at selected layers and reuses them in subsequent
shared layers.
We use the training-free inference mechanism with a fixed uniform schedule, retaining indexers at
$F_{\text{DeepSeek-V3.2}}=\{0,1,5,9,\ldots,57\}$ (16 of 61 layers) and $F_{\text{GLM-5.1}}=\{0,1,5,9,\ldots,77\}$ (21 of 78 layers). All remaining layers reuse the indices from the nearest preceding full layer. The schedule is specified by \texttt{index\_topk\_freq=4}. We use neither a calibration set nor greedy layer-selection search. 
The final retrieval budget is $K=2048$ and IndexCache is enabled for both prefill and decoding. 

\paragraph{PIVOT.}
  PIVOT~\citep{liu2026pivot} aggregates neighboring queries
  into a proxy to share a full-prefix scan, followed by
  direct index reuse or query-specific refinement.
  Our SGLang adaptation follows the PIVOT-Refine selection
  rule, with prefill and MTP verification group sizes both
  set to $4$, a historical candidate budget of $4096$ tokens,
  and a final retrieval budget of $K=2048$ per query.
  We construct the proxy by mean-pooling the indexing queries and gating weights, using the frozen
  native indexer.
  For a group starting at position $a$, query $t$ is refined
  over the shared historical candidates together with
  positions $\{a,\ldots,t\}$, including the query itself.
  We fall back to native DSA indexing when the history
  contains fewer than $4096$ tokens, groups are incomplete,
  or the group layout is unsupported.
  The method is enabled in all $61$ layers of DeepSeek-V3.2
  and all $78$ layers of GLM-5.1 during prefill and MTP target
  verification.

\end{document}